\documentclass[10pt,twocolumn,letterpaper]{article}

\usepackage[pagenumbers]{cvpr} % To force page numbers, e.g. for an arXiv version

\usepackage[dvipsnames]{xcolor}

\usepackage{bbm}

\definecolor{cvprblue}{rgb}{0.21,0.49,0.74}
\usepackage[pagebackref,breaklinks,colorlinks,citecolor=cvprblue]{hyperref}
\usepackage{tikz}
\usepackage{amsmath}
\usepackage{tabularx}
\usepackage{multirow}
\usepackage{colortbl}

\DeclareMathOperator*{\argmax}{\arg\!\max}

\def\confName{CVPR}
\def\confYear{2024}

\title{Would Deep Generative Models Amplify Bias in Future Models?}

\author{Tianwei Chen$^{1*}$, Yusuke Hirota$^{1}$, Mayu Otani$^{2}$, Noa Garcia$^{1}$, Yuta Nakashima$^{1}$ \\
Osaka University$^{1}$,\quad CyberAgent Inc.$^{2}$\\
{\tt\small \{chentw@is., y-hirota@is., noagarcia@, n-yuta@\}ids.osaka-u.ac.jp, otani\_mayu@cyberagent.co.jp}\\
}

\begin{document}
\maketitle

\def\thefootnote{*}\footnotetext{Work done during internship at CyberAgent Inc.}\def\thefootnote{\arabic{footnote}}

\begin{abstract}
We investigate the impact of deep generative models on potential social biases in upcoming computer vision models. As the internet witnesses an increasing influx of AI-generated images, concerns arise regarding inherent biases that may accompany them, potentially leading to the dissemination of harmful content. This paper explores whether a detrimental feedback loop, resulting in bias amplification, would occur if generated images were used as the training data for future models. We conduct simulations by progressively substituting original images in COCO and CC3M datasets with images generated through Stable Diffusion. The modified datasets are used to train OpenCLIP and image captioning models, which we evaluate in terms of quality and bias. Contrary to expectations, our findings indicate that introducing generated images during training does not uniformly amplify bias. Instead, instances of bias mitigation across specific tasks are observed. We further explore the factors that may influence these phenomena, such as artifacts in image generation (\eg, blurry faces) or pre-existing biases in the original datasets.
\end{abstract}    
\section{Introduction}
\label{sec:intro}

\begin{figure}
  \centering
  \includegraphics[width = 0.49\textwidth]{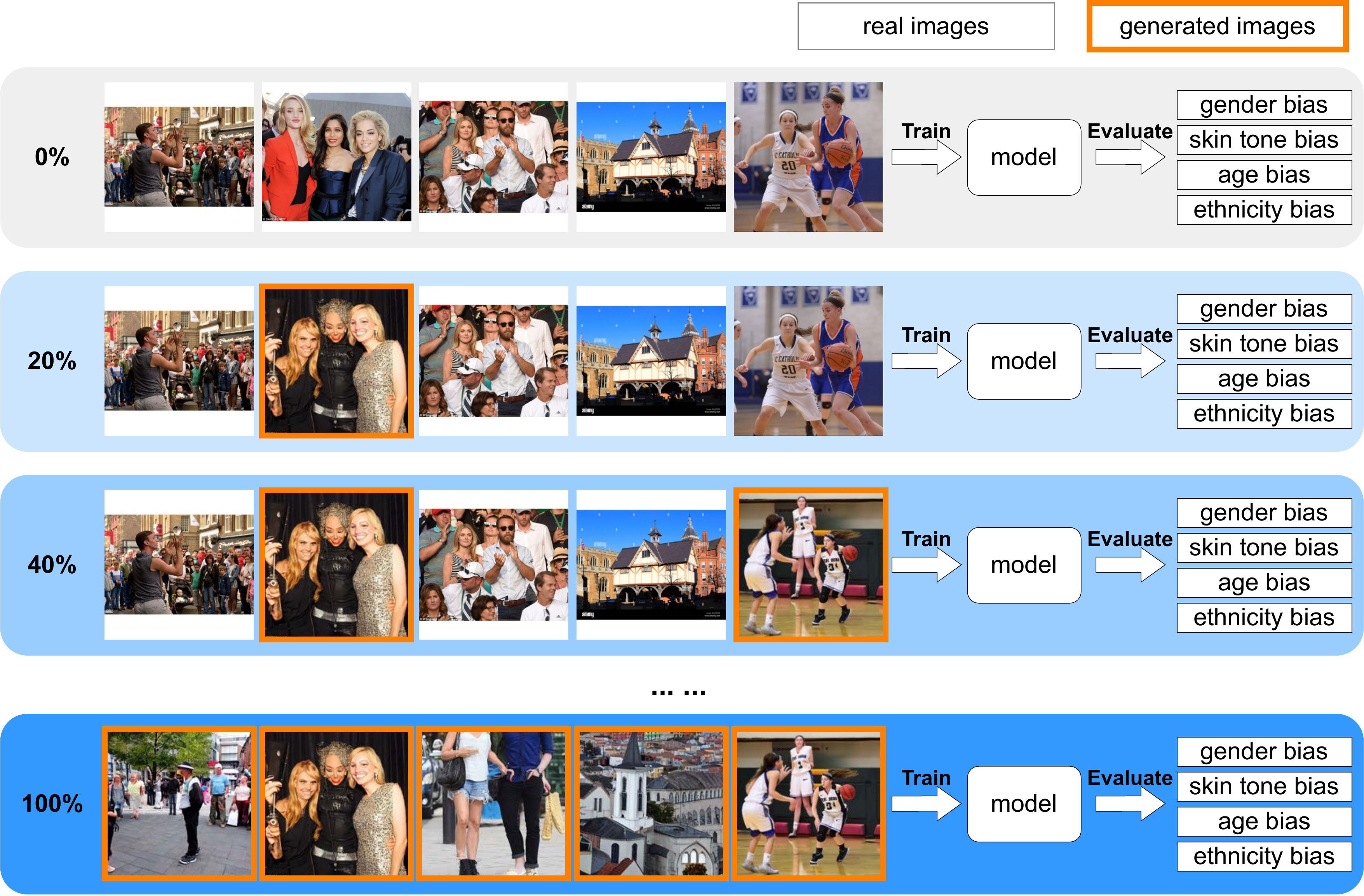}
  \caption{We investigate social biases in the training iterations of future models by simulating scenarios where generated images progressively replace real images in the training data.}
  \label{fig:intro}
\end{figure}

Emerging deep generative models, such as DALL-E 2~\cite{dalle2}, Imagen~\cite{imagen}, or Stable Diffusion~\cite{ldm}, have shown remarkable capabilities in producing high-quality images. Trained on extensive datasets gathered from the internet \cite{cc3m,cc12m,laion400m,laion5b}, these models can generate visually compelling images based on user-customized text inputs or prompts, sparking a surge of enthusiasm for image generation across the online community. However, concerns regarding social biases have been systematically identified \cite{IGMBiasInLifeCycle}, including gender bias~\cite{easilyacce,DallEval,MCSA,AuditingGP,StableBias,T2IAT,BAinT2I,fairSD,FairnessFinetuning,SDExpose, BAParadox}, ethnicity bias~\cite{easilyacce,DallEval,StableBias,SocBiasTT2I,FairnessFinetuning}, and geographical bias~\cite{easilyacce,ExploitingCulBias,SocBiasTT2I,ITTGeoBias}. In particular, previous work~\cite{easilyacce,DallEval,MCSA,SDExpose} has highlighted the tendency of deep generative models to produce biased images even when prompted with ostensibly neutral inputs, uncovering unfair associations between specific social groups and certain attributes~\cite{AuditingGP,StableBias,T2IAT,BAParadox}. A common example is the generation of images depicting occupations, such as doctors and nurses, which have been shown to be strongly tied to gender and race. 

Issues with bias tend to be attributed to the composition of the training data. Training images are frequently scraped from the internet with minimal efforts to filter out problematic samples and address representational disparities. Moreover, in the current context, generated images are continuously shared online and mixed with real images, which means that future computer vision models may inadvertently incorporate large portions of synthetically generated images into their training processes. Coupled with the increasing concerns about the presence of social bias in deep generative models, this raises the following question: \textit{What consequences might arise if images generated by biased models become increasingly involved in the training process of future models?}

To address this question, we conduct experiments focusing on vision-and-language (VL) tasks within a scenario where generated images are progressively integrated into the training data. Specifically, we generate new images for COCO \cite{coco} and CC3M\footnote{CC3M is also known as Google Conceptual Captions or GCC.}~\cite{cc3m} datasets using Stable Diffusion~\cite{ldm}, and we gradually replace the original images in the datasets with their generated counterparts. Our evaluation covers four types of demographic bias -- gender, ethnicity, age, and skin tone -- across two tasks: image-text pretraining and image captioning. 
For image-text pre-training, we evaluate the bias introduced by OpenCLIP~\cite{openclip} on two downstream tasks, \ie image retrieval~\cite{phase,coco_bias} and face attribute recognition~\cite{markedness}. For image captioning, we evaluate the performance of ClipCap~\cite{clipcap} and Transformer~\cite{transformer} using bias metrics such as leakage (LIC)~\cite{LIC} and gender misprediction (Error)~\cite{womenalso,mitigatingG}. 

Our experiments show that the behaviors of the evaluated biases are inconsistent and vary as we gradually replace original images with generated ones. In some cases, biases increase, while in others, they decrease. To understand this phenomenon further, we hypothesize two potential causes: 1) as existing datasets inherently contain biases \cite{coco_bias,phase}, if the bias introduced by the generated images aligns with the pre-existing biases in the dataset, it may not aggravate the existing bias, and 2) artifacts in Stable Diffusion's generations, particularly concerning the generation of human faces (\eg, blurred or poorly defined attributes), may lead models trained on such data to avoid learning demographic features. Overall, the key contributions of this paper are:
\begin{enumerate}
\item We show that, under our experimental setup, generated images from current deep generative models do not consistently amplify bias. Our experiments reveal different levels of bias for gender, ethnicity, age, and skin tone on both the COCO and CC3M datasets when increasing the number of generated images. 
\item Through a set of follow-up experiments, we explore the underlying reasons behind these results, offering valuable insights into the dynamics between image generation models and existing datasets. 
\item We propose recommendations for handling biased generated images in the training process of future models, contributing to the ongoing discourse on responsible and unbiased AI development.
\end{enumerate}

While bias is not consistently amplified in our experiments, we find the presence of bias amplification in multiple instances concerning. Moreover, as our experiments are conducted on moderate-scale datasets with about $3$ million images, representing about $130$ times less data than the original CLIP~\cite{CLIP}, the impact of generated images on large-scale training remains uncertain. We believe that, as a community, addressing bias and ensuring models are safe for everyone should be a top priority. We hope our findings contribute to increased awareness of fairness in computer vision and inspire the creation of models with unbiased and equitable representations.
\section{Related work}

\paragraph{Bias in pre-trained vision-and-language models} 
Pre-trained VL models are not only used in downstream tasks through fine-tuning \cite{12in1, Oscar, VinVL} but also in guiding model training \cite{ldm, LAFITE, PragmaticInferenceWCLIP} and serving as evaluation metrics \cite{hessel2021clipscore, ZeroTextCap, LAFITE}. With the proliferation of VL models, there is an increasing awareness about the inherent biases present in them \cite{WBW,VLStereo,markedness,debiasVL,phase}. 
For example, Wolfe \etal~\cite{markedness} evaluated the proximity of neutral text (\eg, ``a photo of a person'') and an attributive text (\eg, ``a photo of a white person'') in the CLIP embedding space \cite{CLIP}. The differences between demographic groups served as indicators of biases in the models. Chuang \etal~\cite{debiasVL} and Garcia \etal~\cite{phase} explored performance gaps among demographic attributes (\eg, \texttt{\small{man}} and \texttt{\small{woman}} for gender, and \texttt{\small{lighter}} and \texttt{\small{darker}} for skin tone) in downstream tasks, such as classification and image retrieval. Overall, previous work \cite{VL-EAT, markedness, debiasVL, phase} has provided methodologies for detecting and evaluating bias in pre-trained VL models, especially in relation to gender and ethnicity. We leverage these approaches to anticipate potential bias in forthcoming datasets, particularly in scenarios where generated images dominate a significant portion of the online image sources, which is a plausible but underexplored scenario.

\vspace{-10pt}
\paragraph{Synthetic data and pre-trained models}
Synthetically generated data is increasingly influencing the pre-training and fine-tuning processes of VL models, whether intentionally or unintentionally. On the one hand, synthetic data is used as an additional training resource when the original dataset is insufficient \cite{ExpandSC,freeATM,WuNG23} or unreliable \cite{stableRep}.
On the other hand, the widespread dissemination of synthetic images on the internet can inadvertently contaminate datasets~\cite{IGMBiasInLifeCycle}. 
Taori \etal~\cite{DataFeedback} explored the data feedback loop and found that incorporating generated data into subsequent model training rounds could exacerbate dataset biases. Furthermore, Hataya~\etal\cite{corruptingData} showed that models trained on large portions of synthetic data dropped their performance.  Building upon these insights, we study the repercussions of synthetic data on social bias in VL models.
\section{Dataset contamination process}

VL models are trained on pairs of images and text. The process for collecting this type of data typically begins with scraping the internet to gather a set of images $\mathcal{X} = \{x\}$, where $x$ is an image. For smaller or moderately sized datasets~\cite{coco,flickr30k,vgqa}, textual descriptions $y$ for each image $x$ are manually generated by crowdsourcing or in-house annotators, resulting in the set $\mathcal{Y} = \{y\}$. However, for large-scale datasets~\cite{cc3m,cc12m,laion400m,laion5b}, where generating specific annotations is unfeasible, text accompanying the images in the original websites is used, often from the ALT\footnote{ALT text refers to the text in the ALT attribute of HTML tags. 
} text. Subsequently, some form of filtering is applied to remove inappropriate content. Formally, let $p_\mathcal{I}(x)$ and $p_\mathcal{T}(y)$ represent the distributions of collected images and corresponding descriptions. All $x \in \mathcal{X}$ and $y \in \mathcal{Y}$ can be seen as samples from $p_\mathcal{I}(x)$ and $p_\mathcal{T}(y)$, respectively. The textual description $y$ is derived from $x \sim p_\mathcal{I}(x)$ through a framing process $y = f(x)$, which determines what aspects of $x$ to describe. 

Biases in the dataset-creation process are introduced from three main sources~\cite{BiasVisualSurvey}. Firstly, biases are inherited from the original population of images on the internet,\footnote{If the scraping is random sampling, the population is identical to $p_\mathcal{I}(x)$, but typically this is not the case because of filtering.} in which content from specific demographic groups and geographical regions is overrepresented. Secondly, additional biases are introduced by the image descriptions provided by annotators or website authors, reflecting their stereotypes. Lastly, the filtering process itself can introduce additional bias; for instance, in the CC3M dataset, entities appearing less than $100$ times were filtered out, potentially removing content from minority groups. 

We define dataset contamination with generated images (hereafter referred to as \textit{dataset contamination}) as a dataset wherein part of its population is replaced with generated images. That is, someone uploads to the internet images $x' = g(y')$ generated by a generative model $g$ with a prompt $y'$. In this process, we operate under two assumptions: (1) a mental image $\bar{x}$ that people aim to achieve with a generative model also conforms to the distribution $p_\mathcal{I}(x)$, and (2) the image description process from the mental image $\bar{x}$ to a prompt $y'$ has the same framing and bias as $f$. Given these assumptions, we infer that $y'$ adheres to the distribution $p_\mathcal{T}(y)$ as $y' = f(\bar{x})$ and $\bar{x} \sim p_\mathcal{I}(x)$. Therefore, the distribution $p_{\mathcal{G}}(x)$ of generated images is given by:
\begin{align}
    p_\mathcal{G}(x) = \sum_{y} p_{\mathcal{T}\rightarrow \mathcal{G}}(x|y)p_\mathcal{T}(y),
\end{align}
where $p_{\mathcal{T}\rightarrow \mathcal{G}}(x|y)$ corresponds to the generative process $g(y)$. This means that we can generate images from descriptions $y \in \mathcal{Y}$ as described in \cite{corruptingData}. Eventually, we create a dataset $\mathcal{D}(\alpha)$ by sampling images $x$ with a prior $\alpha$ from:
\begin{align}
    \mathcal{D}(\alpha) = \{x \sim (1-\alpha) p_\mathcal{I}(x) + \alpha p_\mathcal{G}(x)\}.
\end{align}

This process of dataset contamination allows us to evaluate the impact of the generative model while keeping the other sources of bias consistent with the original dataset.

\section{Bias evaluation tasks}
\label{sec:preliminary}

The range of tasks in the scope of VL is extensive and diverse. For a survey, please refer to \cite{mogadala2021trends,zhang2024vision}. In this work, we examine the effects of dataset contamination on two fundamental tasks: image-text pre-training and image captioning. Next, we outline bias evaluation in each of them.

\subsection{Image-text pretraining}
\label{sec:bias_pretraining}
Image-text pertaining involves training a model to learn semantic correspondences between visual appearance and text, such as associating the word ``rabbit" with and image of a rabbit. Models like CLIP \cite{CLIP} and its variants \cite{openclip,slip,declip,lit,laclip} are trained on large-scale image-text pairs sourced from the internet. CLIP-like models are reported to exhibit social biases, including gender~\cite{VL-EAT,VLZSGenderBias,debiasVL,markedness,phase}, ethnicity~\cite{debiasVL,markedness,phase}, age~\cite{markedness,phase}, and skin tone~\cite{phase}, and are susceptible to additional biases introduced by dataset contamination. We use OpenCLIP \cite{openclip}, an open-source variant, and assess its performance on text-to-image retrieval, self-similarity, and person preference. 

\vspace{-10pt}
\paragraph{Text-to-image retrieval}
Following Garcia \etal \cite{phase}, where CLIP was shown to perform differently for different demographic attributes (\eg images of men showed a higher recall at $k$ (R$@k$) than images of women), we evaluate text-to-image retrieval performance. Text-to-image retrieval consists on finding the corresponding image given an input text. We compute R$@k$ for different demographic attributes on PHASE \cite{phase} and COCO \cite{coco} datasets for OpenCLIP models trained on datasets $\mathcal{D}(\alpha)$.

\vspace{-10pt}
\paragraph{Self-similarity} Proposed by Wolfe \etal~\cite{markedness}, \textit{self-similarity} evaluates how images of an attribute group are distributed in the embedding space. The core idea is that if a CLIP-like model is trained on numerous images of a specific group with diverse descriptions in the contrastive training process, its encoders will attempt to distribute these images within a larger volume in the embedding space to differentiate them. Otherwise, images of an underrepresented group may occupy a smaller volume. 

Formally, let $\mathcal{E}_a \subset \mathcal{E}$ denote the subset of the entire test set $\mathcal{E}$, containing only samples of a certain attribute group $a$. Self-similarity $\text{SS}(\mathcal{E}_a)$ for group $a$ is given by:
\begin{align}
    \text{SS}(\mathcal{E}_a) = \frac{1}{|\mathcal{E}_a|^2-|\mathcal{E}_a|}\sum_{x, x'} c(x,x'),
\end{align}
where $|\mathcal{E}_a|$ gives the number of samples in $\mathcal{E}_a$, $c(x, x')$ denotes the cosine similarity between $x$ and $x'$ in the embedding space,\footnote{Letting $e_\text{V}$ denote the CLIP visual encoder, $c(x, x')$ is defined as $c(x,x') = \cos(e_\text{V}(x), e_\text{V}(x'))$ where $\cos$ gives the cosine similarity.} and the summation is computed over all combinations of two samples $x$ and $x'$ in $\mathcal{E}_a$. A higher self-similarity means images in $\mathcal{E}_a$ are concentrated in the embedding space.

Different treatments of attribute groups appear in the difference of $\text{SS}(\mathcal{E}_a)$'s among $a$ in attribute $\mathcal{A}$.\footnote{For instance, the binarized gender attribute in PHASE \cite{phase} is given by $\mathcal{A} = \{\texttt{male}, \texttt{female}\}$.}
Self-similarity is defined over the learned embedding space, and the samples in that space give different distributions for different datasets; therefore, self-similarity cannot be compared across models. As we are interested in how broad the distribution for $a \in \mathcal{A}$ are in comparison with others in $\mathcal{A}$, we normalize self-similarity scores as: 
\begin{align}
    \bar{\text{SS}}(\mathcal{E}_a)=\frac{\text{SS}(\mathcal{E}_a)}{\sum_{a \in \mathcal{A}} \text{SS}(\mathcal{E}_a) / |\mathcal{E}_a|} - 1.
\end{align}

\vspace{-10pt}
\paragraph{Person preference}
Another possible reflection of bias in the embedding space is whether a neutral description of an image represents images of a specific attribute group, \ie, if a certain group is well-represented in a dataset, a neutral description may cover the attribute group. \textit{Person preference}~\cite{markedness} evaluates this skew by comparing the similarities among a neutral description (\eg, ``a photo of a \underline{person}''),  a description with a specific attribute group (\eg, ``a photo of a \underline{white person}''), and images of the group. Formally, let $t_\text{N}$ and $t_a$ denote the neutral description and one attributed by $a$. The person preference score over $\mathcal{E}_a$ is given by:
\begin{align}
    \text{PP}(\mathcal{E}_a) =\frac{1}{|\mathcal{E}_a|} \sum_{x \in \mathcal{E}_a} \mathbbm{1}[c(x, t_\text{N}) > c(x, t_a)]
\end{align}
where $\mathbbm{1}$ is the indicator function, and we abuse notation $c$ to represent the cosine similarity between an image and a description, embedding them with appropriate encoders.

\subsection{Image captioning}

Image captioning is the task of generating descriptions for an input image. Descriptions generated by image captioning models~\cite{Oscar,clipcap} have been found to reproduce bias, especially concerning gender and skin-tone \cite{LIC,coco_bias,mitigatingG}. We assess image captioning models trained on data contamination in terms on caption quality, LIC, and gender misprediction.

\vspace{-10pt}
\paragraph{Caption quality} 
Several automatic metrics have been proposed for evaluating captions quality, including BLEU \cite{bleu}, ROUGE \cite{rouge}, METEOR \cite{meteor}, CIDEr \cite{cider}, and SPICE \cite{spice}, 
which mainly involve a lexical comparison between the generated caption and the correspondent ground-truth caption. Alternatively, CLIPScore~\cite{hessel2021clipscore} evaluates the fidelity of a generated caption to the original image. In our experiments, we adopt BLEU-4, CIDEr, SPICE, and CLIPScore. 

\vspace{-10pt}
\paragraph{LIC}

To evaluate social bias amplification in image captioning models, Hirota \etal~\cite{LIC} proposed LIC. This metric evaluates whether the generated captions are more biased than the captions in the original trained dataset. For LIC, a set of captions is assumed to be biased if a protected attribute can be predicted without being explicitly mentioned. 
Specifically, an attribute classifier $h_a(y)$, which gives the likeliness of an attribute group $a$ from a caption $y$, is trained on a training set $\mathcal{C}_\text{T} = \{(y, a)\}$, where $a$ is the ground-truth attribute group. All attribute-specific words\footnote{We use the same list of attribute-specific words as \cite{LIC}.} in the caption $y$ are masked so that the prediction is not trivial. Then, given a validation set $\mathcal{C}_\text{V}$, again with all attribute-specific words being masked, the model's leakage score is computed as:
\begin{align}
    \text{LIC}_\text{M} = \frac{1}{|\mathcal{C}_\text{E}|} \sum_{(y, a) \in \mathcal{C}_\text{E}} h_a(y) \mathbbm{1}[\argmax_{a'} h_{a'}(y) = a]
\end{align}
$\text{LIC}_\text{M}$ gives a higher value if the attribute group is correctly predicted with a higher confidence value even for the masked captions in $\mathcal{C}_\text{E}$, suggesting that the attribute group can be easily predicted from captions. 

The leakage score is also computed for the captions in the original dataset, \ie, $\text{LIC}_\text{D}$ for $\mathcal{Y}$. The final amplification metric $\text{LIC}$ is defined as the difference between the dataset and the model leakage as:
\begin{align}
    \text{LIC} = \text{LIC}_\text{M} - \text{LIC}_\text{D}.
\end{align}

\vspace{-10pt}
\paragraph{Gender misprediction}
Another bias evaluation metric for image captioning is the \textit{Gender missprediction} or \textit{Error}~\cite{womenalso,mitigatingG}, which measures gender mispredictions in the generated captions as:
\begin{align}
    \text{Error} = \frac{N}{M},
\end{align}
where $M$ is the number of generated captions, and $N$ is the number of captions among the $M$ generated captions whose gender group is incorrectly predicted. Gender is considered incorrectly predicted if it contains any words in the attribute-specific word list for the gender opposite to the ground truth gender. For example, for the ground-truth group \texttt{man}, the gender in the generated caption is considered correct if there are no words from the \texttt{woman}-specific word list, such as \textit{girl}.

\begin{figure}
\centering
\begin{subfigure}{.222\textwidth}
  % \centering
  \includegraphics[width=\linewidth]{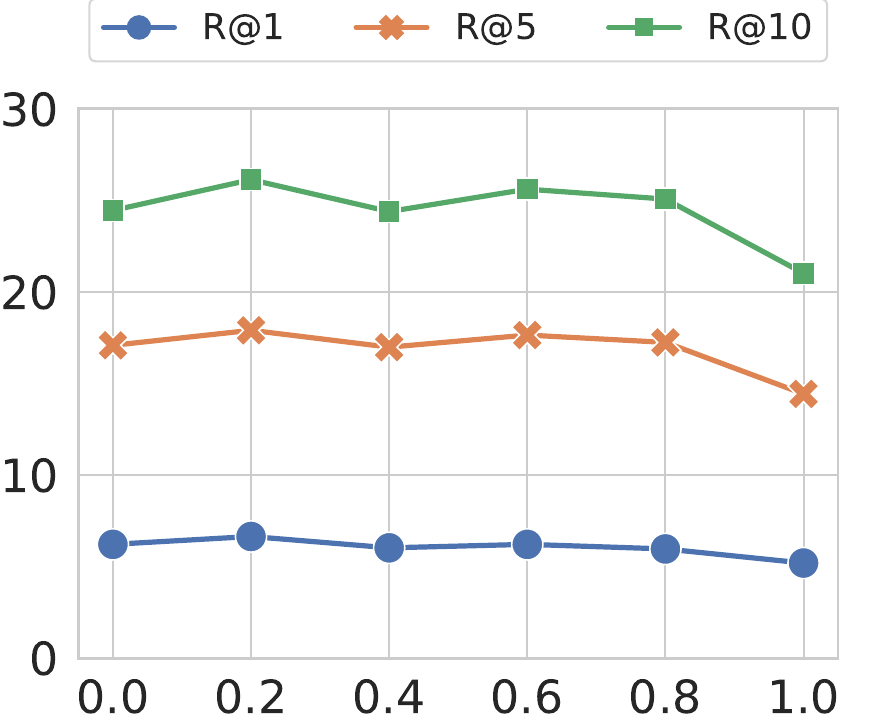}
  \caption{COCO 2014 test set}
  \label{fig:cocoir}
\end{subfigure} \hspace{0.6mm}
\begin{subfigure}{.222\textwidth}
  % \centering
  \includegraphics[width=\linewidth]{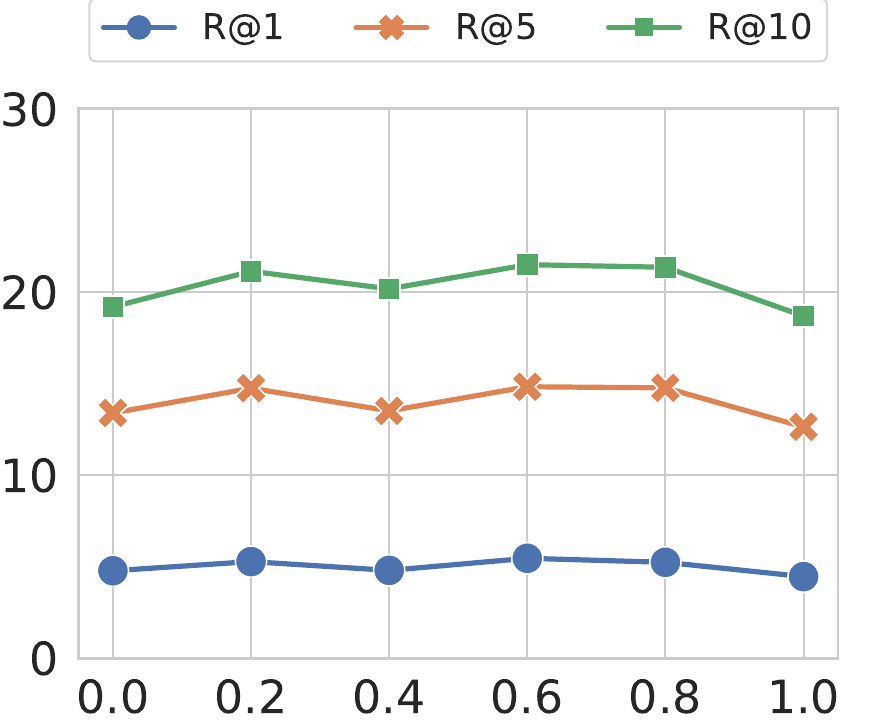}
  \caption{Flickr30K test set}
  \label{fig:flickrir}
\end{subfigure}
\caption{Image retrieval results on COCO 2014 test set and Flickr30k test set for different $\alpha$. The performance of OpenCLIP remains consistent across different levels of dataset contamination.}
\label{fig:nonbias_ir_results}
\end{figure}

\begin{figure*}
% \hspace*{0.5in}
\centering
\resizebox{.999\linewidth}{!}{
\begin{subfigure}{.24\textwidth}
  \centering
  \includegraphics[width=\linewidth]{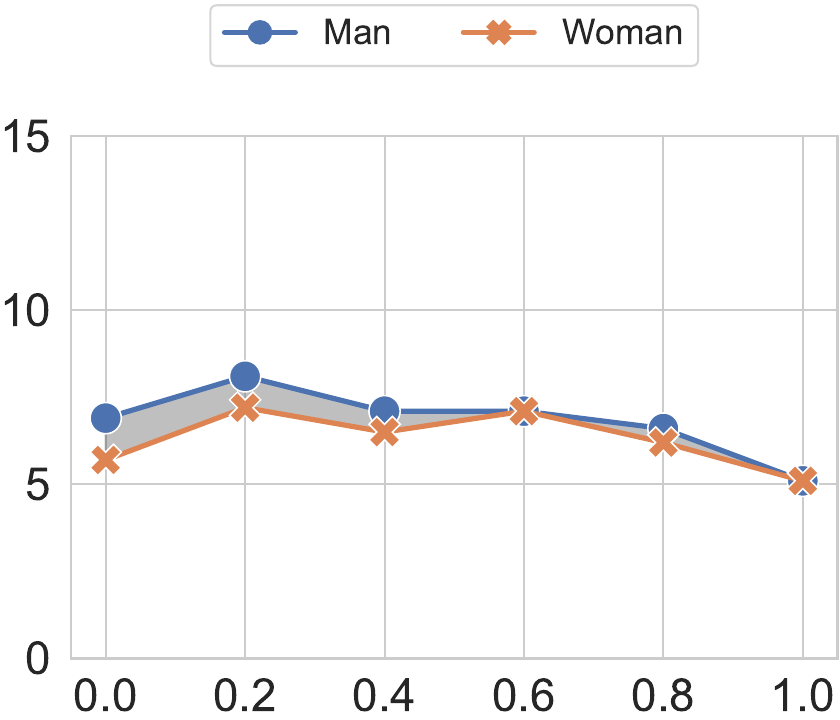}
  \caption{gender}
  \label{fig:phase_gender}
\end{subfigure} % \hspace{0.4mm}
\begin{subfigure}{.24\textwidth}
  % \centering
  \includegraphics[width=\linewidth]{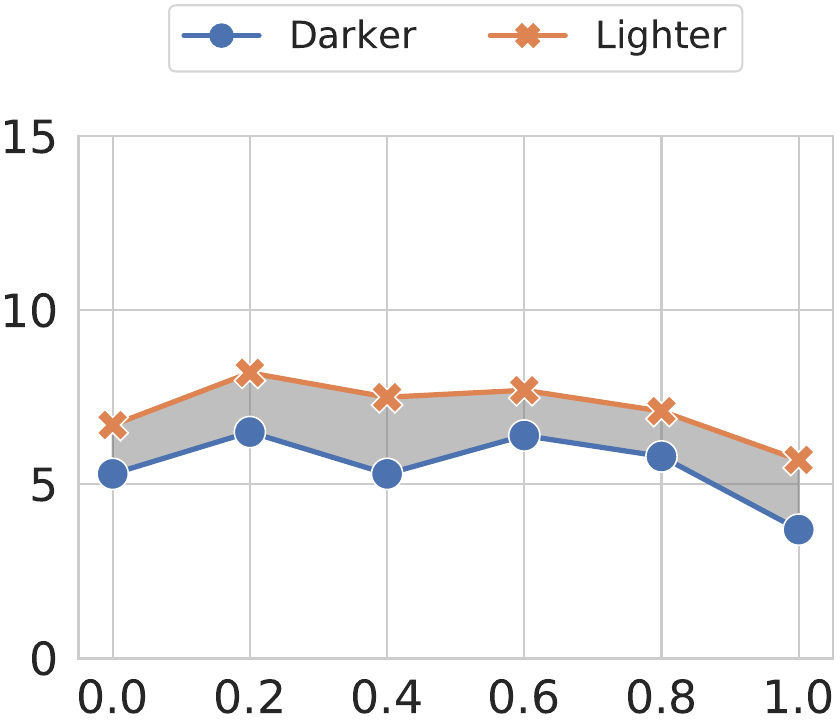}
  \caption{skin tone}
  \label{fig:phase_skintone}
\end{subfigure} % \hspace{0.4mm}
% \hspace*{0.5in}
\begin{subfigure}{.24\textwidth}
  % \centering
  \includegraphics[width=\linewidth]{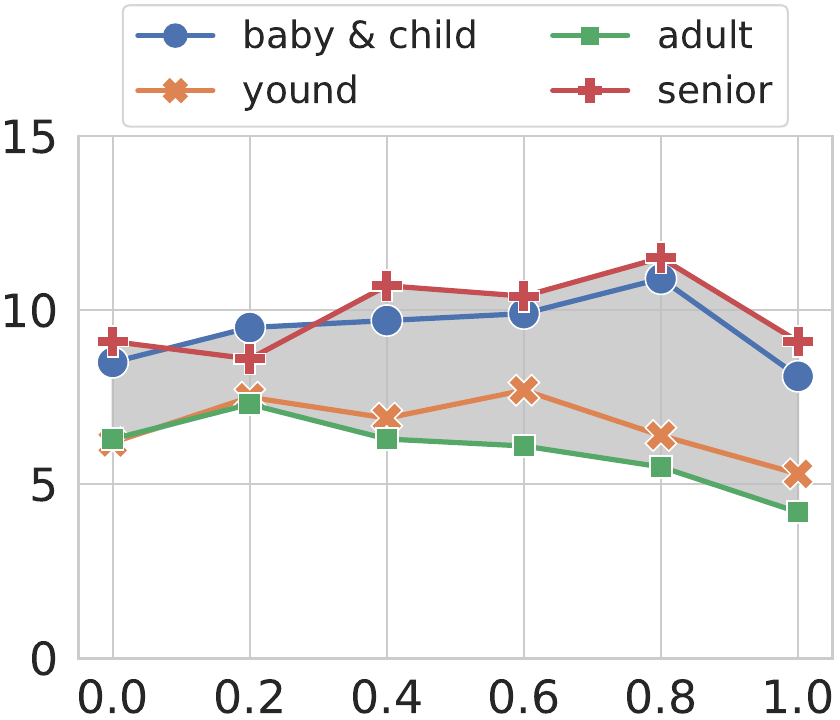}
  \caption{age}
  \label{fig:phase_age}
\end{subfigure} % \hspace{0.25mm}
\begin{subfigure}{.24\textwidth}
  % \centering
  \includegraphics[width=\linewidth]{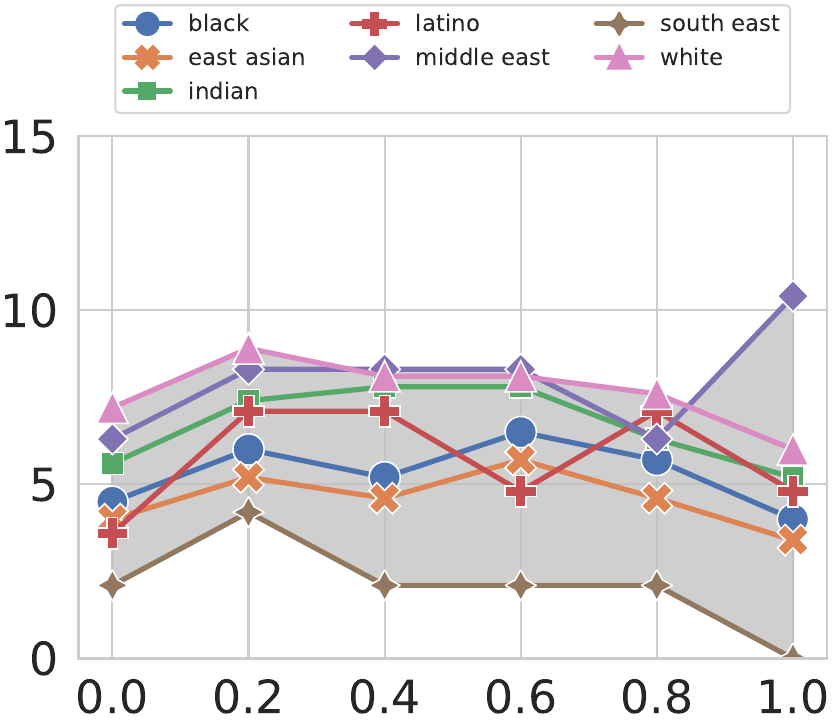}
  \caption{ethnicity}
  \label{fig:phase_ethnicity}
\end{subfigure}
}
\caption{$R@5$ on CC3M using PHASE annotations for different $\alpha$. Bias is highlighted in gray as the difference between groups. We observe different trends: bias mitigation in \cref{fig:phase_gender}, consistency in \cref{fig:phase_skintone}, amplification in \cref{fig:phase_age}, and no clear trend in \cref{fig:phase_ethnicity}.}
\label{fig:phase_results}
\vspace{1mm}
\end{figure*}
\begin{figure}
\centering
\resizebox{1.00\linewidth}{!}{
\begin{subfigure}{.235\textwidth}
  % \centering
  \includegraphics[width=\linewidth]{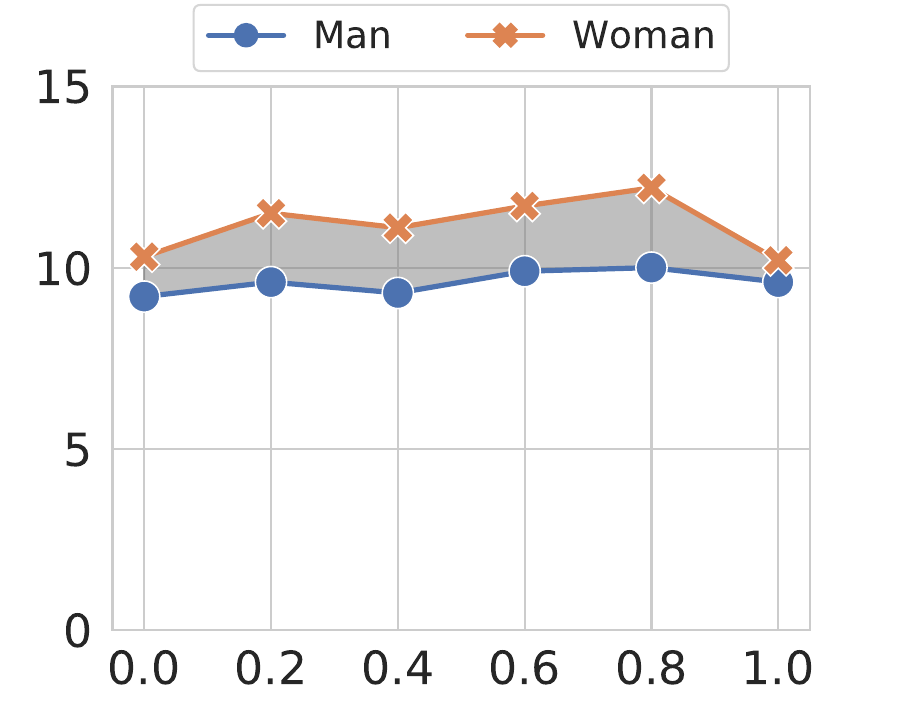}
  \caption{gender}
  \label{fig:cocobias_gender}
\end{subfigure} % \hspace{0.6mm}
\begin{subfigure}{.235\textwidth}
  % \centering
  \includegraphics[width=\linewidth]{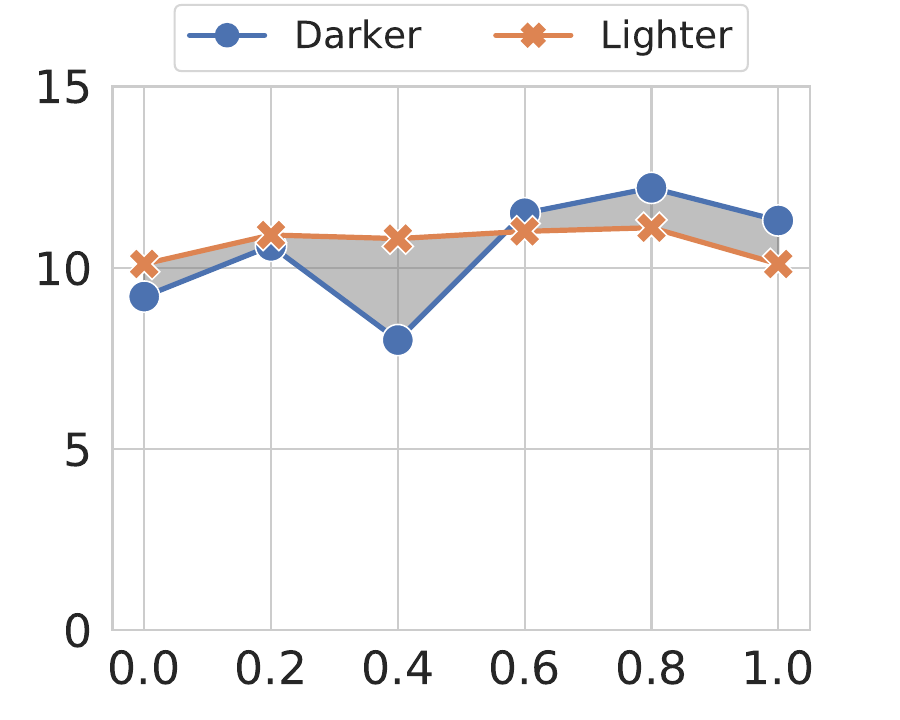}
  \caption{skin tone}
  \label{fig:cocobias_skintone}
\end{subfigure}
}
\caption{$R@5$ on COCO 2014 test set for different $\alpha$. Bias is highlighted in gray as the difference between groups. Both gender and skin tone bias show ambiguous trends.}
\label{fig:cocobias_results}
\end{figure}
\begin{figure*}
\centering
\resizebox{.96\linewidth}{!}{
\begin{subfigure}{.32\textwidth}
  % \centering
  \includegraphics[width=0.95\linewidth]{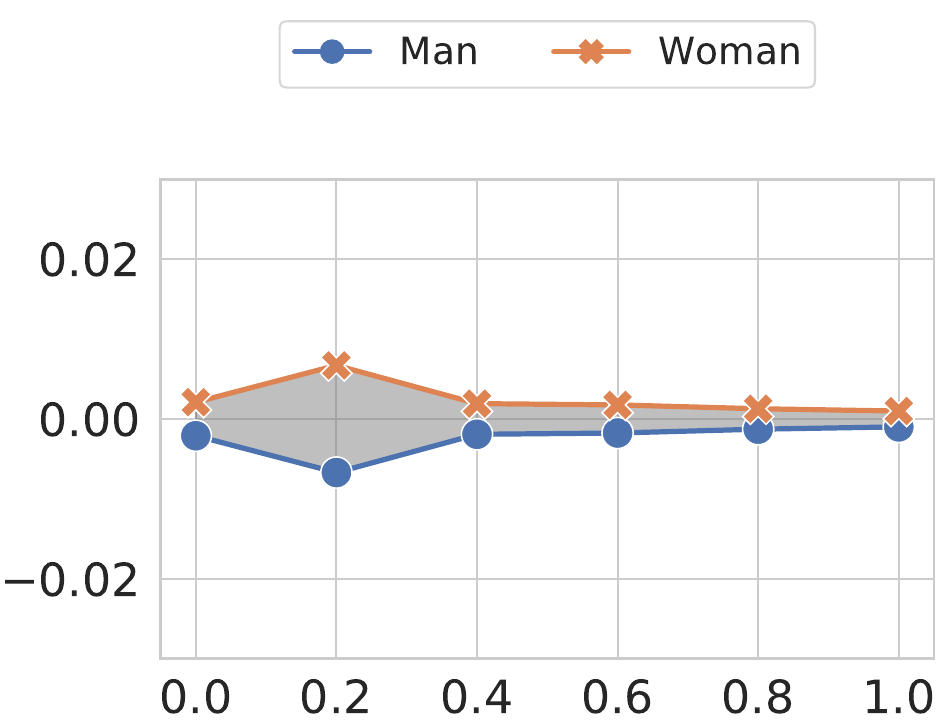}
  \caption{gender}
  \label{fig:self_similarity_gender}
\end{subfigure} % \hspace{4mm}
\begin{subfigure}{.32\textwidth}
  % \centering
  \includegraphics[width=0.95\linewidth]{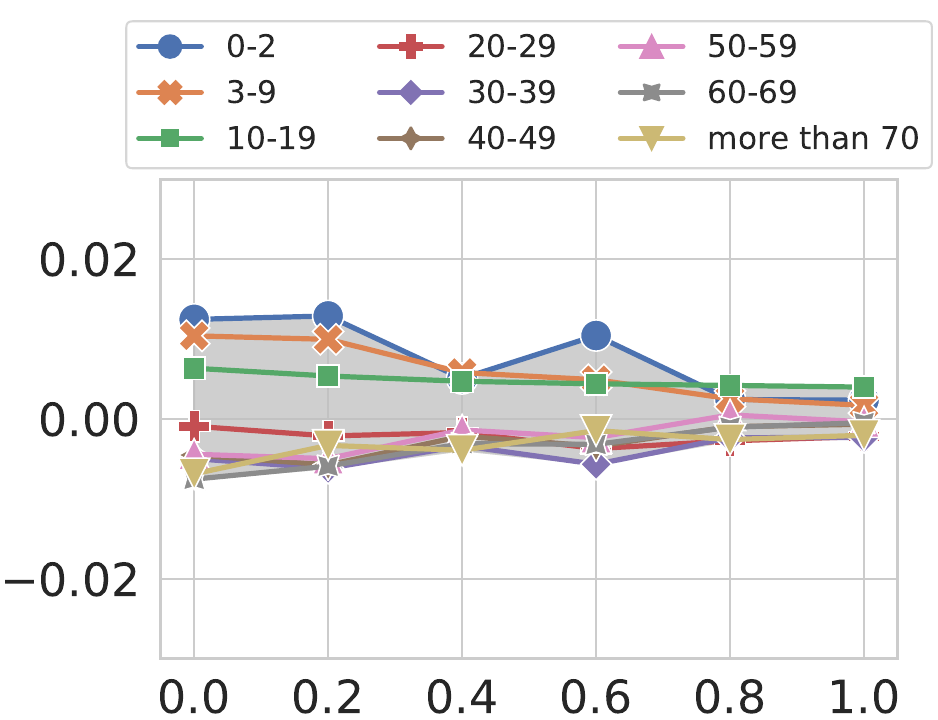}
  \caption{age}
  \label{fig:self_similarity_age}
\end{subfigure}
\begin{subfigure}{.32\textwidth}
  % \centering
  \includegraphics[width=0.95\linewidth]{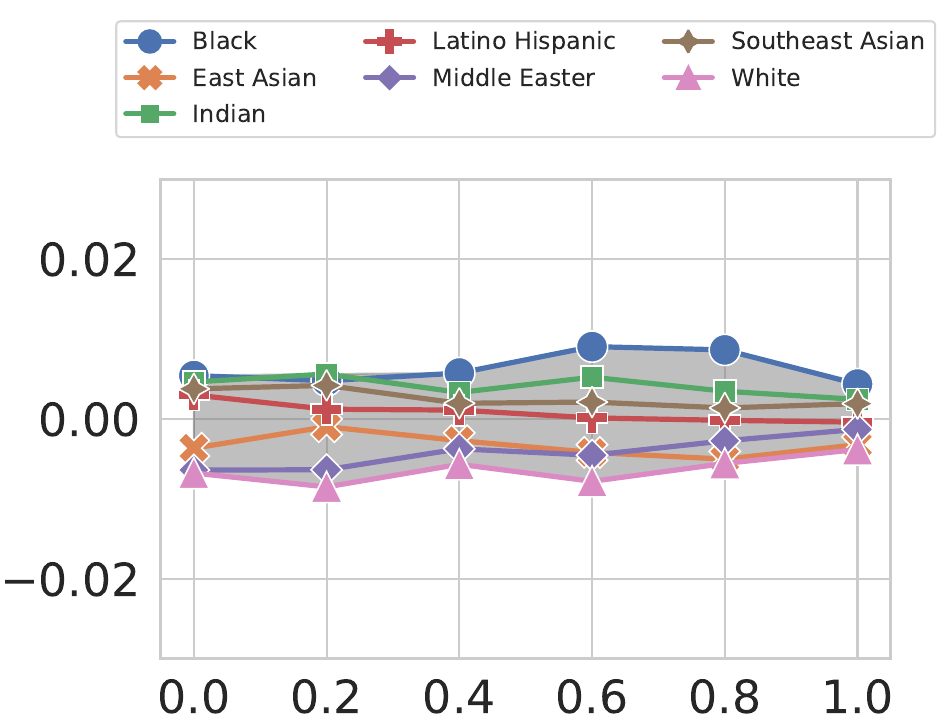}
  \caption{ethnicity}
  \label{fig:self_similarity_ethnicity}
\end{subfigure}
}
\caption{Self-similarity score of each group in the FairFace dataset for different $\alpha$. Bias is highlighted in gray.}
\label{fig:self_similarity_results}
\vspace{2mm}
\end{figure*}
\begin{figure*}
\centering
\resizebox{1.00\linewidth}{!}{
\hspace{8mm}
\begin{subfigure}{.31\textwidth}
  % \centering
  \includegraphics[width=0.9\linewidth]{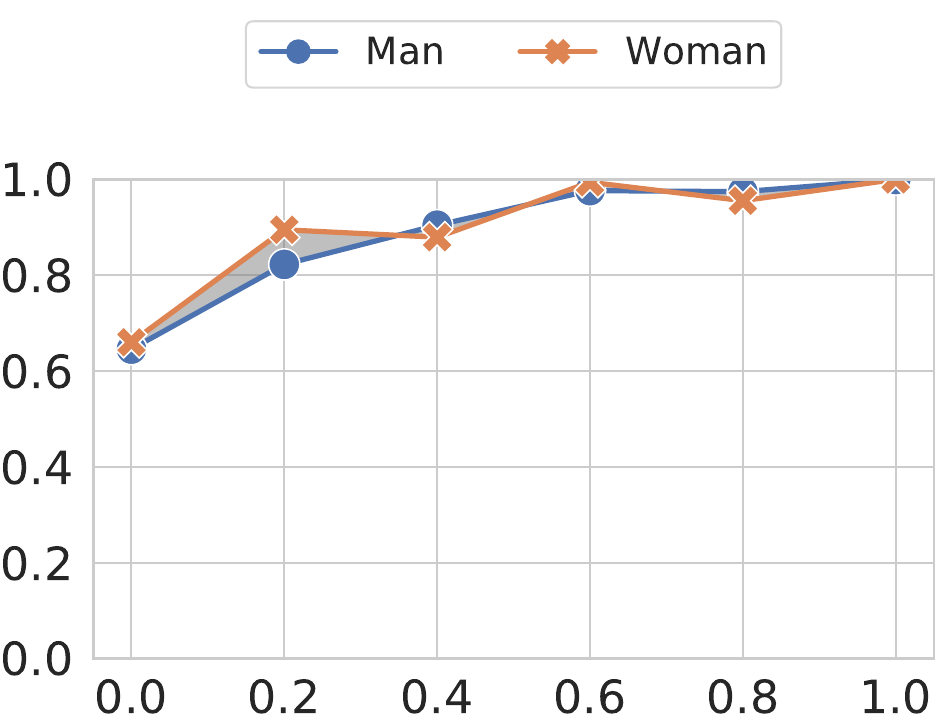}
  \caption{gender}
  \label{fig:person_preference_gender}
\end{subfigure} % \hspace{0.5mm}
\begin{subfigure}{.31\textwidth}
  % \centering
  \includegraphics[width=0.9\linewidth]{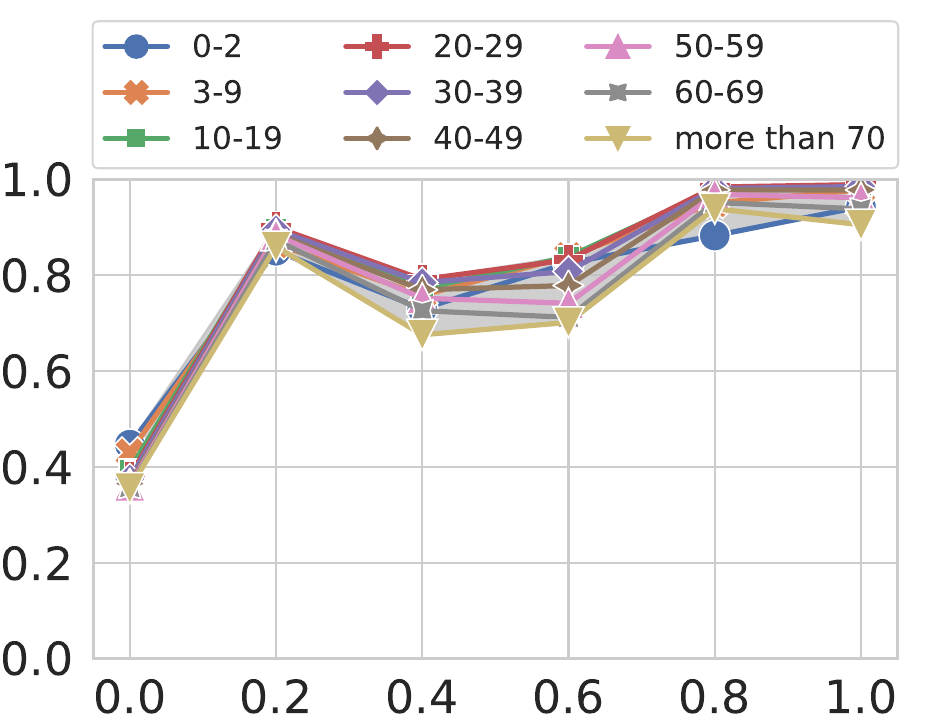}
  \caption{age}
  \label{fig:person_preference_age}
\end{subfigure} % \hspace{0.5mm}
\begin{subfigure}{.31\textwidth}
  % \centering
  \includegraphics[width=0.9\linewidth]{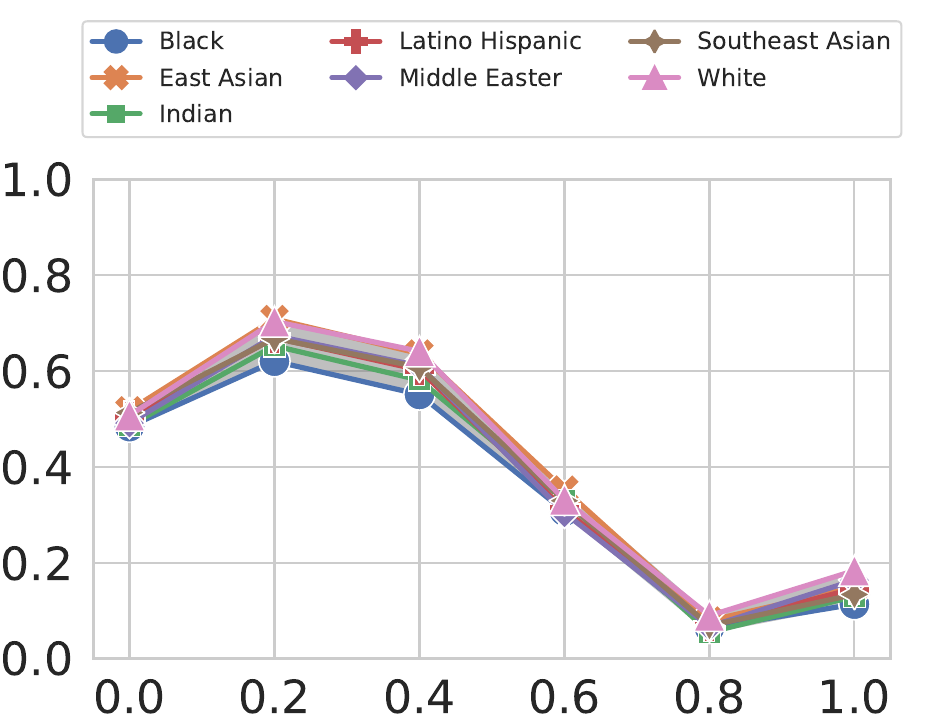}
  \caption{ethnicity}
  \label{fig:person_preference_ethnicity}
\end{subfigure}
}
\caption{Person preference score of each group in the FairFace dataset for different $\alpha$. Bias is highlighted in gray. None of the three figures show a clear tendency. Besides, the changes in bias are relatively small compared with the person preference scores.}
\label{fig:person_preference_results}
\vspace{2mm}
\end{figure*}

\section{Results on OpenCLIP}
\label{sec:exp_openclip}
We train OpenCLIP~\cite{openclip} using various versions of the CC3M~\cite{cc3m} dataset, each with different levels of dataset contamination. For dataset contamination, we use Stable Diffusion v1.5~\cite{ldm} to generate images using the original captions as prompts. Due to the nature of the CC3M dataset, where images are provided as URL links and many of these links have expired, we are only able to retrieve $2,772,289$ valid images for our training data. Consequently, we generate images solely for the prompts corresponding to the available images. We randomly replace $20\%$, $40\%$, $60\%$, $80\%$, and $100\%$ of original images with the images we generate, \ie $\mathcal{D}(\alpha)$ for $\alpha = 0.0$ (the original CC3M dataset), $0.2$, $0.4$, $0.6$, $0.8$, and $1$. 
Evaluation is conducted on five datasets, two for performance evaluation and three for bias evaluation. For performance evaluation, we use the COCO 2014 1K test set~\cite{coco} and the Flickr30k test set~\cite{flickr30k}. For bias evaluation, we use the CC3M validation set using PHASE demographic annotations~\cite{phase}, the COCO validation set using gender and skin-tone annotations~\cite{coco_bias},  and the whole FairFace dataset~\cite{fairface}. We run all experiments three times with different random seeds and report the average.

\subsection{OpenCLIP performance}
We first evaluate the performance of OpenCLIP trained under our experimental settings on two standard datasets: the COCO 2014 test set and the Flickr30K test set. We report text-to-image retrieval performance as R@$k$ with $k=1, 5, 10$. Results are shown in \Cref{fig:nonbias_ir_results}, from which we observe that:

\begin{itemize}
    \item Image retrieval results remain relatively constant for all levels of dataset contamination, from $\mathcal{D}(0.0)$ to $\mathcal{D}(1.0)$, in both datasets and for R@1, R@5, and R@10.
    \item Our reported results on OpenCLIP are considerably lower than those of the original CLIP. We attribute this difference to the disparity in the size of the training set. While our training is conducted with less than $3$ million image-text pairs, the original CLIP model is trained on about $400$ million samples. 
\end{itemize}

In summary, the use of generated images for training OpenCLIP on the CC3M dataset appears to have minimal influence on the retrieval performance of its encoders. Next, we proceed to evaluate the impact of dataset contamination on the bias metrics.

\subsection{Bias in OpenCLIP}
As described in \Cref{sec:bias_pretraining}, text-to-image pertaining bias is evaluated on three metrics: text-to-image retrieval, self-similarity, and person preference. For text-to-image retrieval, we report results on the CC3M validation set with age, gender, skin-tone and ethnicity annotations from PHASE~\cite{phase} (\Cref{fig:phase_results}) and the COCO validation set with gender and skin-tone annotations from \cite{coco_bias} (\Cref{fig:cocobias_results}). For self-similarity and person preference, we report results on the FairFace dataset (\Cref{fig:self_similarity_results,fig:person_preference_results}). From these results, we find the following trends with respect to bias:

\begin{itemize}
    \item \textbf{Consistent bias amplification}: We observe instances of consistent bias amplification, as illustrated in \Cref{fig:phase_age}, where the text-to-image performance gap between the different age groups widens with increasing levels of dataset contamination.
    \item \textbf{Consistent bias mitigation}:  In \Cref{fig:phase_gender,fig:self_similarity_gender}, we observe instances of consistent bias mitigation, where the gender gap is reduced for both text-to-image performance and self-similarity metrics. The gap in self-similarity for the age attribute is also consistently reduced, as shown in \Cref{fig:self_similarity_age}, indicating a bias mitigation effect with the increase of the dataset contamination parameter $\alpha$.
    \item \textbf{Unaffected bias}: In some cases, bias remains unchanged. This is observed in \Cref{fig:phase_skintone}, where the gap in text-to-image retrieval performance between lighter and darker-skin tone images remains constant for the different values of $\alpha$ from $0.0$ to $1.0$.
    \item \textbf{Ambiguous bias trends}: Across most instances, we do not discern a clear bias trend. In \Cref{fig:phase_gender,fig:phase_ethnicity,fig:cocobias_skintone,fig:self_similarity_ethnicity,fig:person_preference_gender,fig:person_preference_ethnicity}, we find no consistent pattern of bias changes, representing half of our experimental results. Unlike \textit{unaffected bias}, the bias in these six experiments fluctuates, showing alternating increases and decreases. For instance, in \Cref{fig:person_preference_gender}, both the woman and man groups intermittently achieve the highest person preference scores. This suggests that multiple factors contribute to bias changes: some amplify bias, while others mitigate it, making bias changes unstable.
\end{itemize}

It is worth noting that the person preference scores show substantial variations in different experiments, surpassing $0.9$ in gender and age (\Cref{fig:person_preference_gender,fig:person_preference_age}), while dropping to $0.2$ for ethnicity (\Cref{fig:person_preference_ethnicity}), despite the unclear trend of bias changes. This observation may be attributed to potential challenges associated with the generation of facial images with Stable Diffusion.

\section{Results on image captioning}
\label{sec:exp_ic}

To analyze bias behavior in image captioning models trained with dataset contamination, we consider two models: Transformer\footnote{Transformer refers to a captioning model with a Transformer-based encoder-decoder where the encoder is ViT-B16 \cite{dosovitskiy2020image}, and the decoder is BERT-base \cite{devlin2018bert}.}~\cite{transformer} and ClipCap \cite{clipcap}. Each model is trained on the COCO 2014 train set~\cite{coco} with different levels of dataset contamination, ranging from $\mathcal{D}(0.0)$ to $\mathcal{D}(1.0)$. Evaluation is conducted in terms of caption quality and bias on the original COCO validation set using gender and skin-tone annotations from \cite{coco_bias}. 

\subsection{Image captioning performance} 
Image captioning results are presented in \Cref{tab:captioning}. Observing the image quality metrics (\ie, BLEU-4, CIDEr, SPICE, and CLIPScore) we note the following: 

\begin{table*}[ht!]
\centering
\caption{Captioning performance and bias metrics for ClipCap and Transformer. }
\renewcommand{\arraystretch}{1.5}
\setlength{\tabcolsep}{4pt}
\resizebox{!}{63pt}{
\begin{tabular}{cccclcccclccclcccc}
\hline 
& \multicolumn{8}{c}{ClipCap} && \multicolumn{8}{c}{Transformer} \\ 
\cline{2-9} \cline{11-18}
& \multicolumn{3}{c}{Bias (↓)} & \multicolumn{1}{c}{} & \multicolumn{4}{c}{Quality (↑)} && \multicolumn{3}{c}{Bias (↓)} & \multicolumn{1}{c}{} & \multicolumn{4}{c}{Quality (↑)} \\ 
\cline{2-4} 
\cline{6-9}
\cline{11-13}
\cline{15-18}
$\alpha$ & LIC-Gender & LIC-Skin & Error  && BLEU-4 & CIDEr & SPICE & CLIPScore  && LIC-Gender & LIC-Skin & Error  && BLEU-4 & CIDEr & SPICE & CLIPScore  \\ \midrule
0  & 3.6  & 1.1  & 5.0 && 31.9 & 105.0  & 20.4  & 76.4  && 3.6 & 2.2 & 11.0 && 28.3 & 92.0 & 18.2 & 72.8 \\
0.2 & 3.8 & 1.9 & 4.7  && 31.8  & 105.1 & 20.4 & 76.8  && 7.6 & 1.6 & 12.1 && 28.4 & 92.1 & 18.0 & 73.1 \\
0.4 & 5.1 & 1.6 & 4.8 && 31.5 & 104.5  & 20.4 & 77.0 && 6.1 & 0.6 & 14.6 && 27.3 & 88.7 & 17.7 & 72.6 \\
0.6 & 3.9 & 1.6 & 4.5 && 31.4 & 104.1 & 20.3 & 77.2 && 5.3 & 2.0 & 10.7 && 26.5 & 88.0 & 17.4 & 73.1 \\
0.8 & 4.1 & 2.0 & 4.6 && 30.7 & 102.4 & 20.0 & 77.4 && 3.9 & 1.9 & 11.1 && 26.8 & 87.7 & 17.3 & 72.8 \\
1.0 & 3.5 & 3.1 & 4.1 && 23.8 & 84.6 & 17.7 & 78.3  && 2.2 & 2.2 & 13.2 && 21.0 & 70.3 & 14.9 & 72.9 \\ \hline
\end{tabular}}
\label{tab:captioning}
\vspace{2mm}
\end{table*}

\begin{itemize}
    \item All lexical similarity-based metrics (\ie, BLUE-4, CIDEr, and SPICE) either experience a gradual decrease or remain relatively stable from $\alpha = 0$, the original dataset, to $0.8$. However, there's a significant drop between $0.8$ and $1.0$, suggesting that even a small amount of real images is necessary to maintain captioning performance.
    \item In contrast, the semantic similarity-based metric (\ie, CLIPScore) remains unaffected by variations in dataset contamination, particularly evident in the case of the Transformer model. While ClipCap slightly improves in CLIPScore, we hypothesize that it is because of the use of CLIP in both image generation and image captioning processes. That is, Stable Diffusion uses CLIP to obtain the text embedding for a caption, so the generated image is strongly tied to it. Therefore, the training set $\mathcal{D}(\alpha)$ with larger $\alpha$ gives image-caption pairs that are close to each other in the CLIP embedding space. ClipCap trained with such a dataset thus only needs to learn the inverse process of the CLIP text encoder, \ie, from an embedding to a caption, for these pairs, which can be easier than learning to fill the gap between images to captions. Thus, ClipCap may easily generate captions that match well with the corresponding images in the CLIP embedding space, consequently increasing CLIPScore. 
\end{itemize}

\subsection{Bias metrics in image captioning} 

With regard to the bias metrics, which include LIC for gender (LIC-gender), LIC for skin-tone (LIC-skin), and gender mispredictions (error), the results are also presented in \Cref{tab:captioning}. We summarize our observations as follows:

\begin{itemize}

    \item \textbf{No trend for gender bias}: LIC scores for gender show no noticeable trend across different values of $\alpha$. In terms of gender mispredictions, similar to the LIC score, there is no clear tendency across the contamination ratios. Under our settings, we cannot draw any definitive conclusion about gender bias. 
    
    \item \textbf{Skin-tone bias amplification}: While LIC for skin-tone on Transformer appears stable, on ClipCap it increases from $1.1$ at $\alpha = 0$ to $3.1$ at $\alpha = 1$. This trend could be attributed to Stable Diffusion accentuating the skin-tone bias present in the original dataset. For example, it has been found that, in the COCO dataset, indoor images tend to feature white people while black people tend to appear indoors \cite{coco_bias}. Similar contextual biases have been observed in Stable Diffusion generations \cite{SocBiasTT2I,easilyacce}.

\end{itemize}
\section{Analysis}
\label{subsec:clip_analysis}

Through our experiments, we observe the existence of different trends in the biases as we progressively replace real images with generated ones. To comprehend the underlying reasons behind this phenomenon, we explore potential factors based on our observations. We primarily focus on two possible explanations: (1) the inherent biases present within the original training datasets, and (2) the limitations of current deep generative models.

\paragraph{Inherent biases in original datasets}
\label{subsec:existing_bias}
Even though Stable Diffusion is known to produce biased images~\cite{easilyacce,DallEval,MCSA,AuditingGP,StableBias,T2IAT,ExploitingCulBias,SocBiasTT2I}, the original datasets, CC3M and COCO datasets, have also been found to be strongly unbalanced~\cite{phase,coco_bias}. 
For example, the CC3M validation set shows large gaps in perceived skin tone, with $3,166$ images of lighter v.s. $318$ images of darker skin-tone people, and perceived ethnicity, with $2,231$ images of White people v.s. $16$ images of Middle Eastern people \cite{phase}.
Similarly, the COCO validation set, has been annotated with $7,466$ images of man v.s. $3,314$ images of woman and $9,873$ images of lighter v.s. $1,096$ images of darker skin-tone people \cite{coco_bias}. If the disparities in representation within the original datasets resemble the biases in the images generated by Stable Diffusion, it is plausible that the biases remain unchanged as real images are progressively replaced with generated ones.

\paragraph{Failure of generation in Stable Diffusion}

\begin{figure}
  \centering
  \includegraphics[width = 0.49\textwidth]{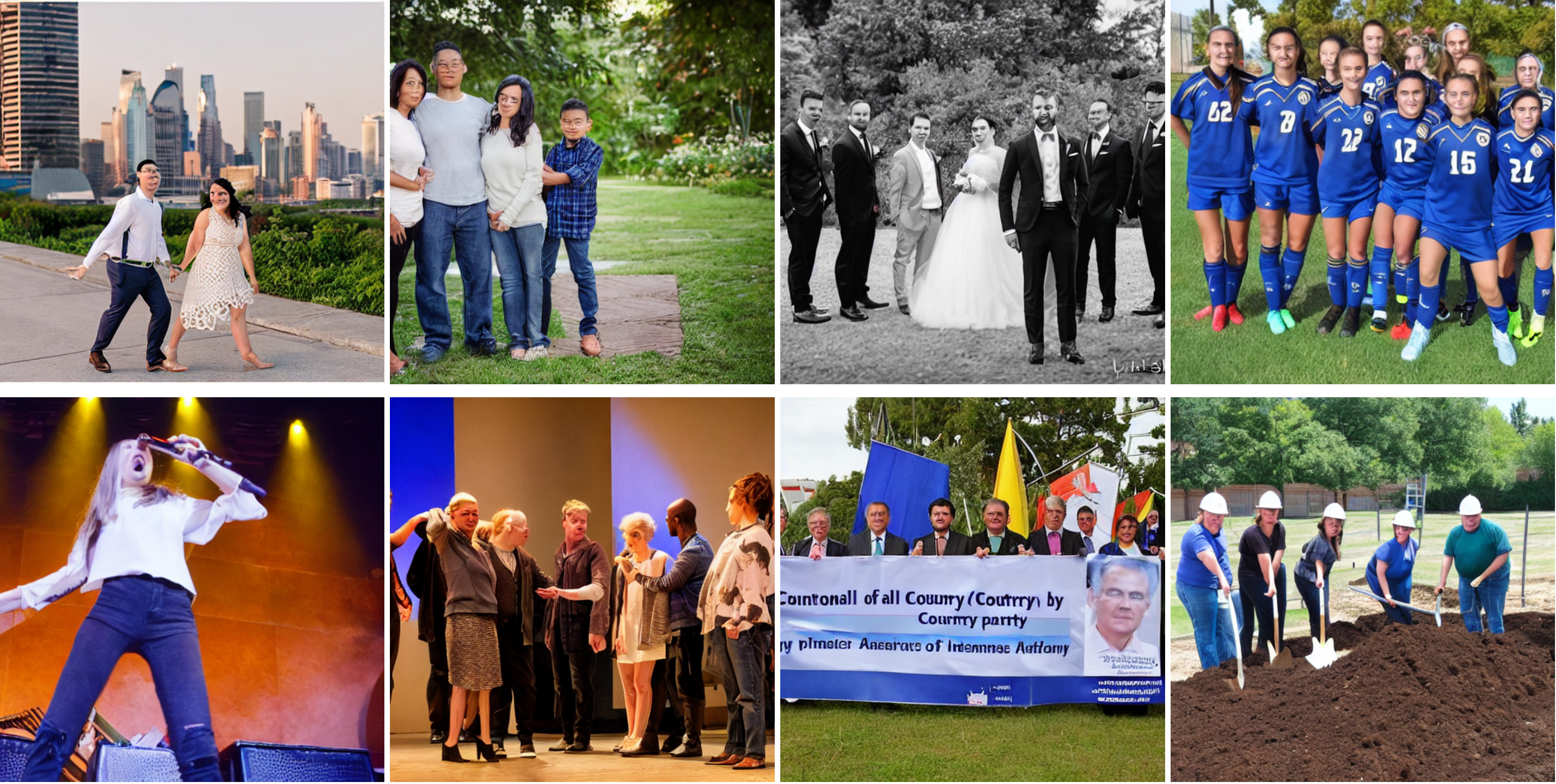}
  \caption{Blurry faces in the generated images. When this happens, the attributes (\eg, gender and age) on the faces are hard to distinguish and further used in the model's training.}
  \label{fig:examples_blurry_face}
  \vspace{2mm}
\end{figure}

\begin{figure}
  \centering
  \includegraphics[width = 0.49\textwidth]{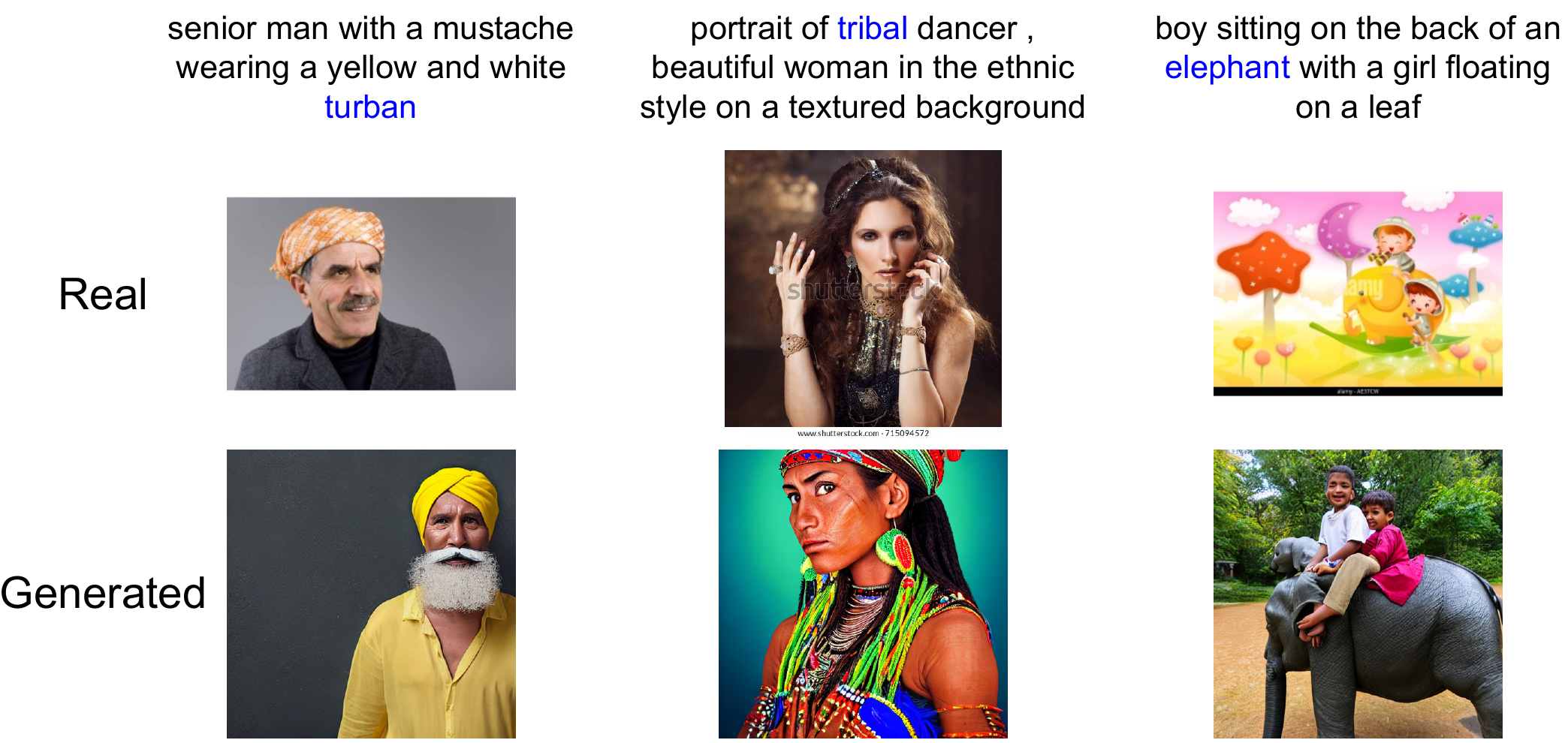}
  \caption{Stereotyping in the generated images. The words in \textcolor{blue}{blue} may cause Stable Diffusion to generate stereotyped images. }
  \label{fig:examples_overreact}
\end{figure}

Deep generative models like Stable Diffusion present several limitations beyond bias concerns. One prominent issue is the tendency for faces to become blurred when generating multiple people. Moreover, Stable Diffusion has been shown to stereotype certain culturally-associated words \cite{ExploitingCulBias}. When examining the generated images in the training dataset, we find similar issues, as shown in  \Cref{fig:examples_blurry_face,fig:examples_overreact}. These issues can impact bias: blurred faces may diminish gender or age biases, while stereotyping could potentially exacerbate ethnicity bias. This phenomenon could elucidate the gender bias mitigation observed in \Cref{fig:phase_gender,fig:self_similarity_gender}.
Overall, due to the complexity of how bias originates and propagates across tasks, there is no one-size-fits-all solution to explain its causes and remedies.

\section{Recommendations}

From our experiments and analysis, we found that while images generated by Stable Diffusion exhibit bias across different demographic attributes, their use for training does not consistently amplify bias. This finding aligns with recent studies~\cite{stableRep,fakeit,syndataimproveclf,laclip} that use generated data from deep generative models for training. These studies highlight the diversity of effects that the generated data can have on model performance, potentially leading to performance improvements. Since the impact of generated data may depend on the original dataset and target task, we propose the following recommendations:

\begin{itemize}
    \item \textbf{Bias-filtering preprocessing}: Considering the possibility that bias in the original dataset could be more pronounced than in deep generative models, we advocate for bias-filtering preprocessing during data collection from the internet, regardless of whether generated images are involved.
\item \textbf{Caution with generation issues}:  While generation issues like blurry faces may aid in bias mitigation in some tasks, they could potentially lead to bias amplification in others. Moreover,  it is important not to regard generation issues as features, as they may be resolved in future iterations of generative models.
\end{itemize}

\section{Limitations}

\begin{itemize}
    \item Due to the scale of current vision-and-language datasets like LAION-400M \cite{laion400m} and LAION-5B \cite{laion5b}, our computational resources are insufficient for generating images and training models on such large datasets. Instead, our experiments are conducted using COCO and CC3M datasets, limiting the scope of insights to be drawn.

   \item The use of Stable Diffusion for image generation may overlook potential findings that could arise from other models with either more biased generations or better bias filtering capabilities.

   \item Our bias evaluation is focused on gender, age, ethnicity, and skin tone. The study does not explore all potential types of bias and leaves out the exploration of intersectional bias, leaving room for further investigation into additional dimensions of bias and fairness.
\end{itemize}

\section{Conclusion}
We investigated the impact of synthetic images generated by Stable Diffusion on bias in future models. We simulated a scenario where the generated images are progressively integrated into future datasets and evaluated bias in two downstream tasks: image-text pertaining with OpenCLIP and image captioning. Our findings revealed that the inclusion of generated images resulted in diverse effects on the downstream tasks, ranging from bias amplification to bias mitigation. Further visualization and analysis provided potential explanations underlying this phenomenon, including the inherent bias in the original datasets and the generation issues associated with Stable Diffusion.

\section{Acknowledgment}
This work is partly supported by JST CREST Grant No. JPMJCR20D3, JST FOREST Grant No. JPMJFR216O, JSPS KAKENHI Nos. JP22K12091 and  JP23H00497.

{
    \small
    \bibliographystyle{ieeenat_fullname}
    \bibliography{main}
}

\maketitlesupplementary

\begin{table}[ht!]
\centering
\caption{Details of the common (with no social attributes) image retrieval results for OpenCLIP.}
\renewcommand{\arraystretch}{1.25}
\begin{subtable}[t]{0.48\textwidth}
\centering
\caption{COCO image retrieval}
\resizebox{!}{25pt}{
\begin{tabular}{ccccccc}
\hline
Flickr30K IR & 0.0 & 0.2 & 0.4 & 0.6 & 0.8 & 1.0 \\ \hline
\rowcolor[HTML]{C0C0C0} 
@1           & 6.23     & 6.66    & 6.04    & 6.23    & 5.98    & 5.21     \\
\rowcolor[HTML]{EFEFEF} 
@5           & 17.09    & 17.91   & 16.99   & 17.65   & 17.25   & 14.44    \\
@10          & 24.45    & 26.13   & 24.39   & 25.61   & 25.07   & 21.02   \\ \hline
\end{tabular}
}
\label{tab:NB_coco}
\end{subtable}

\vspace{6.6mm}

\begin{subtable}[t]{0.48\textwidth}
\centering
\caption{Flickr30k image retrieval}
\resizebox{!}{25pt}{
\begin{tabular}{ccccccc}
\hline
Flickr30K IR & 0.0 & 0.2 & 0.4 & 0.6 & 0.8 & 1.0 \\ \hline
\rowcolor[HTML]{C0C0C0} 
@1           & 4.79     & 5.29    & 4.81    & 5.47    & 5.25    & 4.47     \\
\rowcolor[HTML]{EFEFEF} 
@5           & 13.40    & 14.75   & 13.51   & 14.83   & 14.77   & 12.64    \\
@10          & 19.19    & 21.13   & 20.17   & 21.49   & 21.35   & 18.69   \\ \hline
\end{tabular}
}
\label{tab:NB_flickr}
\end{subtable}
\end{table}

\begin{table}[ht!]
\centering
\caption{Details of the image retrieval performance and bias metrics for OpenCLIP in COCO dataset with gender and skin tone annotation. Characters in the table are: 1) In gender: \textbf{M}an and \textbf{W}oman; 2) In skin tone: \textbf{L}ighter and \textbf{D}arker.}
\renewcommand{\arraystretch}{1.25}
\resizebox{!}{120pt}{
\begin{tabular}{ccccclcc}
\hline
 \multicolumn{1}{l}{}       & \multicolumn{1}{l}{}                              & \multicolumn{1}{l}{}                               & \multicolumn{2}{c}{gender}                                                                                &                          & \multicolumn{2}{c}{skin-tone}                                                                             \\ \cline{4-5} \cline{7-8} 
 \multicolumn{1}{c}{\textbf{$\alpha$}}       & \multicolumn{1}{l}{}                              & \multicolumn{1}{l}{\multirow{-2}{*}{All}}          & M                                                   & W                                                   &                          & L                                                   & D                                                   \\ \hline
                            & \cellcolor[HTML]{C0C0C0}{\color[HTML]{000000} @1} & \cellcolor[HTML]{C0C0C0}{\color[HTML]{000000} 3.1} & \cellcolor[HTML]{C0C0C0}{\color[HTML]{000000} 3.0}  & \cellcolor[HTML]{C0C0C0}{\color[HTML]{000000} 3.2}  & \cellcolor[HTML]{C0C0C0} & \cellcolor[HTML]{C0C0C0}{\color[HTML]{000000} 3.4}  & \cellcolor[HTML]{C0C0C0}{\color[HTML]{000000} 3.2}  \\
                           & \cellcolor[HTML]{EFEFEF}{\color[HTML]{000000} @5} & \cellcolor[HTML]{EFEFEF}{\color[HTML]{000000} 9.2} & \cellcolor[HTML]{EFEFEF}{\color[HTML]{000000} 9.2}  & \cellcolor[HTML]{EFEFEF}{\color[HTML]{000000} 10.3} & \cellcolor[HTML]{EFEFEF} & \cellcolor[HTML]{EFEFEF}{\color[HTML]{000000} 10.1} & \cellcolor[HTML]{EFEFEF}{\color[HTML]{000000} 9.2}  \\
\multirow{-3}{*}{0.0} & @10                                               & 13.7                                               & 13.7                                                & 15.1                                                &                          & 15.0                                                & 13.4                                                \\ \hline
                           & \cellcolor[HTML]{C0C0C0}{\color[HTML]{000000} @1} & \cellcolor[HTML]{C0C0C0}{\color[HTML]{000000} 3.2} & \cellcolor[HTML]{C0C0C0}{\color[HTML]{000000} 2.9}  & \cellcolor[HTML]{C0C0C0}{\color[HTML]{000000} 4.5}  & \cellcolor[HTML]{C0C0C0} & \cellcolor[HTML]{C0C0C0}{\color[HTML]{000000} 3.6}  & \cellcolor[HTML]{C0C0C0}{\color[HTML]{000000} 2.9}  \\
                           & \cellcolor[HTML]{EFEFEF}{\color[HTML]{000000} @5} & \cellcolor[HTML]{EFEFEF}{\color[HTML]{000000} 9.7} & \cellcolor[HTML]{EFEFEF}{\color[HTML]{000000} 9.6}  & \cellcolor[HTML]{EFEFEF}{\color[HTML]{000000} 11.5} & \cellcolor[HTML]{EFEFEF} & \cellcolor[HTML]{EFEFEF}{\color[HTML]{000000} 10.9} & \cellcolor[HTML]{EFEFEF}{\color[HTML]{000000} 10.6} \\
\multirow{-3}{*}{0.2}  & @10                                               & 14.8                                               & 15.1                                                & 17.1                                                &                          & 16.6                                                & 16.5                                                \\ \hline
                           & \cellcolor[HTML]{C0C0C0}{\color[HTML]{000000} @1} & \cellcolor[HTML]{C0C0C0}{\color[HTML]{000000} 3.0} & \cellcolor[HTML]{C0C0C0}{\color[HTML]{000000} 3.1}  & \cellcolor[HTML]{C0C0C0}{\color[HTML]{000000} 3.9}  & \cellcolor[HTML]{C0C0C0} & \cellcolor[HTML]{C0C0C0}{\color[HTML]{000000} 3.7}  & \cellcolor[HTML]{C0C0C0}{\color[HTML]{000000} 1.9}  \\
                           & \cellcolor[HTML]{EFEFEF}{\color[HTML]{000000} @5} & \cellcolor[HTML]{EFEFEF}{\color[HTML]{000000} 9.2} & \cellcolor[HTML]{EFEFEF}{\color[HTML]{000000} 9.3}  & \cellcolor[HTML]{EFEFEF}{\color[HTML]{000000} 11.1} & \cellcolor[HTML]{EFEFEF} & \cellcolor[HTML]{EFEFEF}{\color[HTML]{000000} 10.8} & \cellcolor[HTML]{EFEFEF}{\color[HTML]{000000} 8.0}  \\
\multirow{-3}{*}{0.4}  & @10                                               & 13.8                                               & 14.1                                                & 16.2                                                &                          & 15.8                                                & 14.3                                                \\ \hline
                           & \cellcolor[HTML]{C0C0C0}{\color[HTML]{000000} @1} & \cellcolor[HTML]{C0C0C0}{\color[HTML]{000000} 3.4} & \cellcolor[HTML]{C0C0C0}{\color[HTML]{000000} 3.4}  & \cellcolor[HTML]{C0C0C0}{\color[HTML]{000000} 4.1}  & \cellcolor[HTML]{C0C0C0} & \cellcolor[HTML]{C0C0C0}{\color[HTML]{000000} 3.8}  & \cellcolor[HTML]{C0C0C0}{\color[HTML]{000000} 3.3}  \\
                           & \cellcolor[HTML]{EFEFEF}{\color[HTML]{000000} @5} & \cellcolor[HTML]{EFEFEF}{\color[HTML]{000000} 9.7} & \cellcolor[HTML]{EFEFEF}{\color[HTML]{000000} 9.9}  & \cellcolor[HTML]{EFEFEF}{\color[HTML]{000000} 11.7} & \cellcolor[HTML]{EFEFEF} & \cellcolor[HTML]{EFEFEF}{\color[HTML]{000000} 11.0} & \cellcolor[HTML]{EFEFEF}{\color[HTML]{000000} 11.5} \\
\multirow{-3}{*}{0.6}  & @10                                               & 14.7                                               & 15.1                                                & 17.6                                                &                          & 16.7                                                & 17.1                                                \\ \hline
                           & \cellcolor[HTML]{C0C0C0}{\color[HTML]{000000} @1} & \cellcolor[HTML]{C0C0C0}{\color[HTML]{000000} 3.5} & \cellcolor[HTML]{C0C0C0}{\color[HTML]{000000} 3.4}  & \cellcolor[HTML]{C0C0C0}{\color[HTML]{000000} 4.9}  & \cellcolor[HTML]{C0C0C0} & \cellcolor[HTML]{C0C0C0}{\color[HTML]{000000} 3.9}  & \cellcolor[HTML]{C0C0C0}{\color[HTML]{000000} 4.8}  \\
                           & \cellcolor[HTML]{EFEFEF}{\color[HTML]{000000} @5} & \cellcolor[HTML]{EFEFEF}{\color[HTML]{000000} 9.8} & \cellcolor[HTML]{EFEFEF}{\color[HTML]{000000} 10.0} & \cellcolor[HTML]{EFEFEF}{\color[HTML]{000000} 12.2} & \cellcolor[HTML]{EFEFEF} & \cellcolor[HTML]{EFEFEF}{\color[HTML]{000000} 11.1} & \cellcolor[HTML]{EFEFEF}{\color[HTML]{000000} 12.2} \\
\multirow{-3}{*}{0.8}  & @10                                               & 14.5                                               & 14.9                                                & 17.7                                                &                          & 16.5                                                & 17.1                                                \\ \hline
                           & \cellcolor[HTML]{C0C0C0}{\color[HTML]{000000} @1} & \cellcolor[HTML]{C0C0C0}{\color[HTML]{000000} 2.7} & \cellcolor[HTML]{C0C0C0}{\color[HTML]{000000} 3.0}  & \cellcolor[HTML]{C0C0C0}{\color[HTML]{000000} 3.0}  & \cellcolor[HTML]{C0C0C0} & \cellcolor[HTML]{C0C0C0}{\color[HTML]{000000} 3.2}  & \cellcolor[HTML]{C0C0C0}{\color[HTML]{000000} 3.5}  \\
                           & \cellcolor[HTML]{EFEFEF}{\color[HTML]{000000} @5} & \cellcolor[HTML]{EFEFEF}{\color[HTML]{000000} 8.8} & \cellcolor[HTML]{EFEFEF}{\color[HTML]{000000} 9.6}  & \cellcolor[HTML]{EFEFEF}{\color[HTML]{000000} 10.2} & \cellcolor[HTML]{EFEFEF} & \cellcolor[HTML]{EFEFEF}{\color[HTML]{000000} 10.1} & \cellcolor[HTML]{EFEFEF}{\color[HTML]{000000} 11.3} \\
\multirow{-3}{*}{1.0} & @10                                               & 13.1                                               & 14.4                                                & 15.2                                                &                          & 15.1                                                & 16.2                                                \\ \hline
\end{tabular}
}
\label{tab:cocobias}
\end{table}

\begin{table*}[htp!]
\centering
\captionof{table}{Details of the image retrieval performance and bias metrics for OpenCLIP in PHASE dataset. Characters in each table are: 1) In gender: \textbf{M}an and \textbf{W}oman; 2) In age: \textbf{B}aby and \textbf{C}hild; 3) In skin tone: \textbf{L}ighter and \textbf{D}arker; 4) In ethnicity: \textbf{B}lack, \textbf{E}ast \textbf{A}sia, \textbf{I}ndian, \textbf{L}atino, \textbf{M}iddle \textbf{E}ast, \textbf{S}outheast \textbf{A}sia, and \textbf{Wh}ite.}
\renewcommand{\arraystretch}{1.25}
\resizebox{!}{120pt}{
\begin{tabular}{ccrrrrrlrrlrrlrrrrrrr}
\hline
\multicolumn{1}{l}{}       & \multicolumn{1}{l}{}                              & \multicolumn{1}{l}{}                               & \multicolumn{4}{c}{age}                                                                                                                                                                                             &                          & \multicolumn{2}{c}{gender}                                                                              &                          & \multicolumn{2}{c}{skin-tone}                                                                           &                          & \multicolumn{7}{c}{Ethnicity}                                                                                                                                                                                                                                                                                                                                                     \\ \cline{4-7} \cline{9-10} \cline{12-13} \cline{15-21} 
\multicolumn{1}{c}{$\alpha$}       & \multicolumn{1}{l}{}                              & \multicolumn{1}{l}{\multirow{-2}{*}{All}}          & \multicolumn{1}{c}{B\&C}                            & \multicolumn{1}{c}{young}                          & \multicolumn{1}{c}{adult}                          & \multicolumn{1}{c}{senior}                          &                          & \multicolumn{1}{c}{M}                              & \multicolumn{1}{c}{W}                              &                          & \multicolumn{1}{c}{L}                              & \multicolumn{1}{c}{D}                              &                          & \multicolumn{1}{c}{B}                              & \multicolumn{1}{c}{EA}                             & \multicolumn{1}{c}{I}                              & \multicolumn{1}{c}{L}                              & \multicolumn{1}{c}{ME}                              & \multicolumn{1}{c}{SA}                             & \multicolumn{1}{c}{Wh}                              \\ \hline
                           & \cellcolor[HTML]{C0C0C0}{\color[HTML]{000000} @1} & \cellcolor[HTML]{C0C0C0}{\color[HTML]{000000} 2.3} & \cellcolor[HTML]{C0C0C0}{\color[HTML]{000000} 3.0}  & \cellcolor[HTML]{C0C0C0}{\color[HTML]{000000} 2.3} & \cellcolor[HTML]{C0C0C0}{\color[HTML]{000000} 2.4} & \cellcolor[HTML]{C0C0C0}{\color[HTML]{000000} 3.9}  & \cellcolor[HTML]{C0C0C0} & \cellcolor[HTML]{C0C0C0}{\color[HTML]{000000} 2.6} & \cellcolor[HTML]{C0C0C0}{\color[HTML]{000000} 2.3} & \cellcolor[HTML]{C0C0C0} & \cellcolor[HTML]{C0C0C0}{\color[HTML]{000000} 2.6} & \cellcolor[HTML]{C0C0C0}{\color[HTML]{000000} 1.9} & \cellcolor[HTML]{C0C0C0} & \cellcolor[HTML]{C0C0C0}{\color[HTML]{000000} 1.2} & \cellcolor[HTML]{C0C0C0}{\color[HTML]{000000} 1.7} & \cellcolor[HTML]{C0C0C0}{\color[HTML]{000000} 2.2} & \cellcolor[HTML]{C0C0C0}{\color[HTML]{000000} 1.2} & \cellcolor[HTML]{C0C0C0}{\color[HTML]{000000} 0.0}  & \cellcolor[HTML]{C0C0C0}{\color[HTML]{000000} 0.0} & \cellcolor[HTML]{C0C0C0}{\color[HTML]{000000} 2.8} \\
                           & \cellcolor[HTML]{EFEFEF}{\color[HTML]{000000} @5} & \cellcolor[HTML]{EFEFEF}{\color[HTML]{000000} 6.3} & \cellcolor[HTML]{EFEFEF}{\color[HTML]{000000} 8.5}  & \cellcolor[HTML]{EFEFEF}{\color[HTML]{000000} 6.2} & \cellcolor[HTML]{EFEFEF}{\color[HTML]{000000} 6.3} & \cellcolor[HTML]{EFEFEF}{\color[HTML]{000000} 9.1}  & \cellcolor[HTML]{EFEFEF} & \cellcolor[HTML]{EFEFEF}{\color[HTML]{000000} 6.9} & \cellcolor[HTML]{EFEFEF}{\color[HTML]{000000} 5.7} & \cellcolor[HTML]{EFEFEF} & \cellcolor[HTML]{EFEFEF}{\color[HTML]{000000} 6.7} & \cellcolor[HTML]{EFEFEF}{\color[HTML]{000000} 5.3} & \cellcolor[HTML]{EFEFEF} & \cellcolor[HTML]{EFEFEF}{\color[HTML]{000000} 4.5} & \cellcolor[HTML]{EFEFEF}{\color[HTML]{000000} 4.0} & \cellcolor[HTML]{EFEFEF}{\color[HTML]{000000} 5.6} & \cellcolor[HTML]{EFEFEF}{\color[HTML]{000000} 3.6} & \cellcolor[HTML]{EFEFEF}{\color[HTML]{000000} 6.3}  & \cellcolor[HTML]{EFEFEF}{\color[HTML]{000000} 2.1} & \cellcolor[HTML]{EFEFEF}{\color[HTML]{000000} 7.2} \\
\multirow{-3}{*}{0.0} & @10                                               & 9.1                                                & 11.5                                                & 8.9                                                & 9.1                                                & 12.2                                                &                          & 9.7                                                & 8.6                                                &                          & 9.5                                                & 8.7                                                &                          & 8.2                                                & 7.5                                                & 8.9                                                & 6.0                                                & 6.3                                                 & 4.2                                                & 10.0                                               \\ \hline
                           & \cellcolor[HTML]{C0C0C0}{\color[HTML]{000000} @1} & \cellcolor[HTML]{C0C0C0}{\color[HTML]{000000} 2.8} & \cellcolor[HTML]{C0C0C0}{\color[HTML]{000000} 3.3}  & \cellcolor[HTML]{C0C0C0}{\color[HTML]{000000} 2.7} & \cellcolor[HTML]{C0C0C0}{\color[HTML]{000000} 2.7} & \cellcolor[HTML]{C0C0C0}{\color[HTML]{000000} 5.2}  & \cellcolor[HTML]{C0C0C0} & \cellcolor[HTML]{C0C0C0}{\color[HTML]{000000} 3.1} & \cellcolor[HTML]{C0C0C0}{\color[HTML]{000000} 2.7} & \cellcolor[HTML]{C0C0C0} & \cellcolor[HTML]{C0C0C0}{\color[HTML]{000000} 3.2} & \cellcolor[HTML]{C0C0C0}{\color[HTML]{000000} 1.8} & \cellcolor[HTML]{C0C0C0} & \cellcolor[HTML]{C0C0C0}{\color[HTML]{000000} 1.2} & \cellcolor[HTML]{C0C0C0}{\color[HTML]{000000} 1.7} & \cellcolor[HTML]{C0C0C0}{\color[HTML]{000000} 2.2} & \cellcolor[HTML]{C0C0C0}{\color[HTML]{000000} 3.6} & \cellcolor[HTML]{C0C0C0}{\color[HTML]{000000} 2.1}  & \cellcolor[HTML]{C0C0C0}{\color[HTML]{000000} 2.1} & \cellcolor[HTML]{C0C0C0}{\color[HTML]{000000} 3.5} \\
                           & \cellcolor[HTML]{EFEFEF}{\color[HTML]{000000} @5} & \cellcolor[HTML]{EFEFEF}{\color[HTML]{000000} 7.6} & \cellcolor[HTML]{EFEFEF}{\color[HTML]{000000} 9.5}  & \cellcolor[HTML]{EFEFEF}{\color[HTML]{000000} 7.5} & \cellcolor[HTML]{EFEFEF}{\color[HTML]{000000} 7.3} & \cellcolor[HTML]{EFEFEF}{\color[HTML]{000000} 8.6}  & \cellcolor[HTML]{EFEFEF} & \cellcolor[HTML]{EFEFEF}{\color[HTML]{000000} 8.1} & \cellcolor[HTML]{EFEFEF}{\color[HTML]{000000} 7.2} & \cellcolor[HTML]{EFEFEF} & \cellcolor[HTML]{EFEFEF}{\color[HTML]{000000} 8.2} & \cellcolor[HTML]{EFEFEF}{\color[HTML]{000000} 6.5} & \cellcolor[HTML]{EFEFEF} & \cellcolor[HTML]{EFEFEF}{\color[HTML]{000000} 6.0} & \cellcolor[HTML]{EFEFEF}{\color[HTML]{000000} 5.2} & \cellcolor[HTML]{EFEFEF}{\color[HTML]{000000} 7.4} & \cellcolor[HTML]{EFEFEF}{\color[HTML]{000000} 7.1} & \cellcolor[HTML]{EFEFEF}{\color[HTML]{000000} 8.3}  & \cellcolor[HTML]{EFEFEF}{\color[HTML]{000000} 4.2} & \cellcolor[HTML]{EFEFEF}{\color[HTML]{000000} 8.9} \\
\multirow{-3}{*}{0.2}  & @10                                               & 11.7                                               & 15.6                                                & 11.5                                               & 11.4                                               & 13.8                                                &                          & 12.6                                               & 11.0                                               &                          & 12.4                                               & 11.1                                               &                          & 10.3                                               & 5.2                                                & 13.7                                               & 10.7                                               & 10.4                                                & 6.3                                                & 13.3                                               \\ \hline
                           & \cellcolor[HTML]{C0C0C0}{\color[HTML]{000000} @1} & \cellcolor[HTML]{C0C0C0}{\color[HTML]{000000} 2.4} & \cellcolor[HTML]{C0C0C0}{\color[HTML]{000000} 4.1}  & \cellcolor[HTML]{C0C0C0}{\color[HTML]{000000} 2.4} & \cellcolor[HTML]{C0C0C0}{\color[HTML]{000000} 2.0} & \cellcolor[HTML]{C0C0C0}{\color[HTML]{000000} 4.4}  & \cellcolor[HTML]{C0C0C0} & \cellcolor[HTML]{C0C0C0}{\color[HTML]{000000} 2.7} & \cellcolor[HTML]{C0C0C0}{\color[HTML]{000000} 2.1} & \cellcolor[HTML]{C0C0C0} & \cellcolor[HTML]{C0C0C0}{\color[HTML]{000000} 2.8} & \cellcolor[HTML]{C0C0C0}{\color[HTML]{000000} 1.7} & \cellcolor[HTML]{C0C0C0} & \cellcolor[HTML]{C0C0C0}{\color[HTML]{000000} 1.4} & \cellcolor[HTML]{C0C0C0}{\color[HTML]{000000} 2.3} & \cellcolor[HTML]{C0C0C0}{\color[HTML]{000000} 4.1} & \cellcolor[HTML]{C0C0C0}{\color[HTML]{000000} 2.4} & \cellcolor[HTML]{C0C0C0}{\color[HTML]{000000} 6.3}  & \cellcolor[HTML]{C0C0C0}{\color[HTML]{000000} 2.1} & \cellcolor[HTML]{C0C0C0}{\color[HTML]{000000} 2.9} \\
                           & \cellcolor[HTML]{EFEFEF}{\color[HTML]{000000} @5} & \cellcolor[HTML]{EFEFEF}{\color[HTML]{000000} 6.9} & \cellcolor[HTML]{EFEFEF}{\color[HTML]{000000} 9.7}  & \cellcolor[HTML]{EFEFEF}{\color[HTML]{000000} 6.9} & \cellcolor[HTML]{EFEFEF}{\color[HTML]{000000} 6.3} & \cellcolor[HTML]{EFEFEF}{\color[HTML]{000000} 10.7} & \cellcolor[HTML]{EFEFEF} & \cellcolor[HTML]{EFEFEF}{\color[HTML]{000000} 7.1} & \cellcolor[HTML]{EFEFEF}{\color[HTML]{000000} 6.5} & \cellcolor[HTML]{EFEFEF} & \cellcolor[HTML]{EFEFEF}{\color[HTML]{000000} 7.5} & \cellcolor[HTML]{EFEFEF}{\color[HTML]{000000} 5.3} & \cellcolor[HTML]{EFEFEF} & \cellcolor[HTML]{EFEFEF}{\color[HTML]{000000} 5.2} & \cellcolor[HTML]{EFEFEF}{\color[HTML]{000000} 4.6} & \cellcolor[HTML]{EFEFEF}{\color[HTML]{000000} 7.8} & \cellcolor[HTML]{EFEFEF}{\color[HTML]{000000} 7.1} & \cellcolor[HTML]{EFEFEF}{\color[HTML]{000000} 8.3}  & \cellcolor[HTML]{EFEFEF}{\color[HTML]{000000} 2.1} & \cellcolor[HTML]{EFEFEF}{\color[HTML]{000000} 8.1} \\
\multirow{-3}{*}{0.4}  & @10                                               & 10.3                                               & 13.8                                                & 10.6                                               & 9.3                                                & 14.3                                                &                          & 10.4                                               & 10.1                                               &                          & 11.0                                               & 8.8                                                &                          & 9.1                                                & 5.2                                                & 11.5                                               & 8.3                                                & 18.8                                                & 4.2                                                & 11.7                                               \\ \hline
                           & \cellcolor[HTML]{C0C0C0}{\color[HTML]{000000} @1} & \cellcolor[HTML]{C0C0C0}{\color[HTML]{000000} 2.4} & \cellcolor[HTML]{C0C0C0}{\color[HTML]{000000} 2.9}  & \cellcolor[HTML]{C0C0C0}{\color[HTML]{000000} 2.8} & \cellcolor[HTML]{C0C0C0}{\color[HTML]{000000} 1.9} & \cellcolor[HTML]{C0C0C0}{\color[HTML]{000000} 5.2}  & \cellcolor[HTML]{C0C0C0} & \cellcolor[HTML]{C0C0C0}{\color[HTML]{000000} 2.3} & \cellcolor[HTML]{C0C0C0}{\color[HTML]{000000} 2.5} & \cellcolor[HTML]{C0C0C0} & \cellcolor[HTML]{C0C0C0}{\color[HTML]{000000} 2.7} & \cellcolor[HTML]{C0C0C0}{\color[HTML]{000000} 2.0} & \cellcolor[HTML]{C0C0C0} & \cellcolor[HTML]{C0C0C0}{\color[HTML]{000000} 1.9} & \cellcolor[HTML]{C0C0C0}{\color[HTML]{000000} 1.1} & \cellcolor[HTML]{C0C0C0}{\color[HTML]{000000} 3.3} & \cellcolor[HTML]{C0C0C0}{\color[HTML]{000000} 1.2} & \cellcolor[HTML]{C0C0C0}{\color[HTML]{000000} 2.1}  & \cellcolor[HTML]{C0C0C0}{\color[HTML]{000000} 0.0} & \cellcolor[HTML]{C0C0C0}{\color[HTML]{000000} 2.9} \\
                           & \cellcolor[HTML]{EFEFEF}{\color[HTML]{000000} @5} & \cellcolor[HTML]{EFEFEF}{\color[HTML]{000000} 7.0} & \cellcolor[HTML]{EFEFEF}{\color[HTML]{000000} 9.9}  & \cellcolor[HTML]{EFEFEF}{\color[HTML]{000000} 7.7} & \cellcolor[HTML]{EFEFEF}{\color[HTML]{000000} 6.1} & \cellcolor[HTML]{EFEFEF}{\color[HTML]{000000} 10.4} & \cellcolor[HTML]{EFEFEF} & \cellcolor[HTML]{EFEFEF}{\color[HTML]{000000} 7.1} & \cellcolor[HTML]{EFEFEF}{\color[HTML]{000000} 7.1} & \cellcolor[HTML]{EFEFEF} & \cellcolor[HTML]{EFEFEF}{\color[HTML]{000000} 7.7} & \cellcolor[HTML]{EFEFEF}{\color[HTML]{000000} 6.4} & \cellcolor[HTML]{EFEFEF} & \cellcolor[HTML]{EFEFEF}{\color[HTML]{000000} 6.5} & \cellcolor[HTML]{EFEFEF}{\color[HTML]{000000} 5.7} & \cellcolor[HTML]{EFEFEF}{\color[HTML]{000000} 7.8} & \cellcolor[HTML]{EFEFEF}{\color[HTML]{000000} 4.8} & \cellcolor[HTML]{EFEFEF}{\color[HTML]{000000} 8.3}  & \cellcolor[HTML]{EFEFEF}{\color[HTML]{000000} 2.1} & \cellcolor[HTML]{EFEFEF}{\color[HTML]{000000} 8.1} \\
\multirow{-3}{*}{0.6}  & @10                                               & 10.7                                               & 15.8                                                & 11.1                                               & 9.4                                                & 14.6                                                &                          & 11.0                                               & 10.4                                               &                          & 11.5                                               & 9.4                                                &                          & 9.3                                                & 8.0                                                & 13.7                                               & 8.3                                                & 10.4                                                & 4.2                                                & 12.0                                               \\ \hline
                           & \cellcolor[HTML]{C0C0C0}{\color[HTML]{000000} @1} & \cellcolor[HTML]{C0C0C0}{\color[HTML]{000000} 2.2} & \cellcolor[HTML]{C0C0C0}{\color[HTML]{000000} 4.0}  & \cellcolor[HTML]{C0C0C0}{\color[HTML]{000000} 2.1} & \cellcolor[HTML]{C0C0C0}{\color[HTML]{000000} 1.9} & \cellcolor[HTML]{C0C0C0}{\color[HTML]{000000} 4.4}  & \cellcolor[HTML]{C0C0C0} & \cellcolor[HTML]{C0C0C0}{\color[HTML]{000000} 2.2} & \cellcolor[HTML]{C0C0C0}{\color[HTML]{000000} 2.2} & \cellcolor[HTML]{C0C0C0} & \cellcolor[HTML]{C0C0C0}{\color[HTML]{000000} 2.5} & \cellcolor[HTML]{C0C0C0}{\color[HTML]{000000} 1.3} & \cellcolor[HTML]{C0C0C0} & \cellcolor[HTML]{C0C0C0}{\color[HTML]{000000} 1.4} & \cellcolor[HTML]{C0C0C0}{\color[HTML]{000000} 1.1} & \cellcolor[HTML]{C0C0C0}{\color[HTML]{000000} 2.2} & \cellcolor[HTML]{C0C0C0}{\color[HTML]{000000} 3.6} & \cellcolor[HTML]{C0C0C0}{\color[HTML]{000000} 0.0}  & \cellcolor[HTML]{C0C0C0}{\color[HTML]{000000} 0.0} & \cellcolor[HTML]{C0C0C0}{\color[HTML]{000000} 2.8} \\
                           & \cellcolor[HTML]{EFEFEF}{\color[HTML]{000000} @5} & \cellcolor[HTML]{EFEFEF}{\color[HTML]{000000} 6.5} & \cellcolor[HTML]{EFEFEF}{\color[HTML]{000000} 10.9} & \cellcolor[HTML]{EFEFEF}{\color[HTML]{000000} 6.4} & \cellcolor[HTML]{EFEFEF}{\color[HTML]{000000} 5.5} & \cellcolor[HTML]{EFEFEF}{\color[HTML]{000000} 11.5} & \cellcolor[HTML]{EFEFEF} & \cellcolor[HTML]{EFEFEF}{\color[HTML]{000000} 6.6} & \cellcolor[HTML]{EFEFEF}{\color[HTML]{000000} 6.2} & \cellcolor[HTML]{EFEFEF} & \cellcolor[HTML]{EFEFEF}{\color[HTML]{000000} 7.1} & \cellcolor[HTML]{EFEFEF}{\color[HTML]{000000} 5.8} & \cellcolor[HTML]{EFEFEF} & \cellcolor[HTML]{EFEFEF}{\color[HTML]{000000} 5.7} & \cellcolor[HTML]{EFEFEF}{\color[HTML]{000000} 4.6} & \cellcolor[HTML]{EFEFEF}{\color[HTML]{000000} 6.3} & \cellcolor[HTML]{EFEFEF}{\color[HTML]{000000} 7.1} & \cellcolor[HTML]{EFEFEF}{\color[HTML]{000000} 6.3}  & \cellcolor[HTML]{EFEFEF}{\color[HTML]{000000} 2.1} & \cellcolor[HTML]{EFEFEF}{\color[HTML]{000000} 7.6} \\
\multirow{-3}{*}{0.8}  & @10                                               & 9.9                                                & 15.0                                                & 9.5                                                & 8.9                                                & 15.4                                                &                          & 10.2                                               & 9.4                                                &                          & 10.7                                               & 9.0                                                &                          & 8.1                                                & 7.5                                                & 12.2                                               & 9.5                                                & 10.4                                                & 4.2                                                & 11.2                                               \\ \hline
                           & \cellcolor[HTML]{C0C0C0}{\color[HTML]{000000} @1} & \cellcolor[HTML]{C0C0C0}{\color[HTML]{000000} 1.6} & \cellcolor[HTML]{C0C0C0}{\color[HTML]{000000} 3.1}  & \cellcolor[HTML]{C0C0C0}{\color[HTML]{000000} 1.6} & \cellcolor[HTML]{C0C0C0}{\color[HTML]{000000} 1.3} & \cellcolor[HTML]{C0C0C0}{\color[HTML]{000000} 3.1}  & \cellcolor[HTML]{C0C0C0} & \cellcolor[HTML]{C0C0C0}{\color[HTML]{000000} 1.7} & \cellcolor[HTML]{C0C0C0}{\color[HTML]{000000} 1.7} & \cellcolor[HTML]{C0C0C0} & \cellcolor[HTML]{C0C0C0}{\color[HTML]{000000} 1.9} & \cellcolor[HTML]{C0C0C0}{\color[HTML]{000000} 0.8} & \cellcolor[HTML]{C0C0C0} & \cellcolor[HTML]{C0C0C0}{\color[HTML]{000000} 1.2} & \cellcolor[HTML]{C0C0C0}{\color[HTML]{000000} 0.6} & \cellcolor[HTML]{C0C0C0}{\color[HTML]{000000} 1.5} & \cellcolor[HTML]{C0C0C0}{\color[HTML]{000000} 1.2} & \cellcolor[HTML]{C0C0C0}{\color[HTML]{000000} 4.2}  & \cellcolor[HTML]{C0C0C0}{\color[HTML]{000000} 0.0} & \cellcolor[HTML]{C0C0C0}{\color[HTML]{000000} 2.0} \\
                           & \cellcolor[HTML]{EFEFEF}{\color[HTML]{000000} @5} & \cellcolor[HTML]{EFEFEF}{\color[HTML]{000000} 5.1} & \cellcolor[HTML]{EFEFEF}{\color[HTML]{000000} 8.1}  & \cellcolor[HTML]{EFEFEF}{\color[HTML]{000000} 5.3} & \cellcolor[HTML]{EFEFEF}{\color[HTML]{000000} 4.2} & \cellcolor[HTML]{EFEFEF}{\color[HTML]{000000} 9.1}  & \cellcolor[HTML]{EFEFEF} & \cellcolor[HTML]{EFEFEF}{\color[HTML]{000000} 5.1} & \cellcolor[HTML]{EFEFEF}{\color[HTML]{000000} 5.1} & \cellcolor[HTML]{EFEFEF} & \cellcolor[HTML]{EFEFEF}{\color[HTML]{000000} 5.7} & \cellcolor[HTML]{EFEFEF}{\color[HTML]{000000} 3.7} & \cellcolor[HTML]{EFEFEF} & \cellcolor[HTML]{EFEFEF}{\color[HTML]{000000} 4.0} & \cellcolor[HTML]{EFEFEF}{\color[HTML]{000000} 3.4} & \cellcolor[HTML]{EFEFEF}{\color[HTML]{000000} 5.2} & \cellcolor[HTML]{EFEFEF}{\color[HTML]{000000} 4.8} & \cellcolor[HTML]{EFEFEF}{\color[HTML]{000000} 10.4} & \cellcolor[HTML]{EFEFEF}{\color[HTML]{000000} 0.0} & \cellcolor[HTML]{EFEFEF}{\color[HTML]{000000} 6.0} \\
\multirow{-3}{*}{1.0} & @10                                               & 7.9                                                & 12.7                                                & 7.7                                                & 6.6                                                & 12.5                                                &                          & 8.0                                                & 7.6                                                &                          & 8.6                                                & 6.2                                                &                          & 6.7                                                & 6.3                                                & 8.1                                                & 7.1                                                & 14.6                                                & 2.1                                                & 9.2                                                \\ \hline
\end{tabular}
}
\label{tab:phase}
\end{table*}

\begin{table*}[ht!]
\centering
\caption{Details of the normalized self similarity score for OpenCLIP. Characters in each table are: 1) In gender: \textbf{M}an and \textbf{W}oman; 2) In ethnicity: \textbf{B}lack, \textbf{E}ast \textbf{A}sia, \textbf{I}ndian, \textbf{L}atino, \textbf{M}iddle \textbf{E}ast, \textbf{S}outheast \textbf{A}sia, and \textbf{Wh}ite.}
\renewcommand{\arraystretch}{1.6}
\resizebox{!}{50pt}{
\begin{tabular}{crrrrrrrrrrrrrrrrrrrr}
\hline
\multicolumn{1}{l}{} & \multicolumn{2}{c}{gender}                    & \multicolumn{1}{l}{} & \multicolumn{7}{c}{ethnicity}                                                                                                                                            & \multicolumn{1}{l}{} & \multicolumn{9}{c}{age}                                                                                                                                                                                                                                      \\ \cline{2-3} \cline{5-11} \cline{13-21} 
\multicolumn{1}{c}{\textbf{$\alpha$}} & \multicolumn{1}{c}{M} & \multicolumn{1}{c}{W} & \multicolumn{1}{l}{} & \multicolumn{1}{c}{EA} & \multicolumn{1}{c}{I} & \multicolumn{1}{c}{SA} & \multicolumn{1}{c}{Wh} & \multicolumn{1}{c}{ME} & \multicolumn{1}{c}{L} & \multicolumn{1}{c}{B} & \multicolumn{1}{l}{} & \multicolumn{1}{c}{0-2} & \multicolumn{1}{c}{3-9} & \multicolumn{1}{c}{10-19} & \multicolumn{1}{c}{20-29} & \multicolumn{1}{c}{30-39} & \multicolumn{1}{c}{40-49} & \multicolumn{1}{c}{50-59} & \multicolumn{1}{c}{60-69} & \multicolumn{1}{c}{more than 70} \\ \hline
0.0             & -0.210                & 0.210                 &                      & -0.366                 & 0.463                 & 0.377                  & -0.678                & -0.640                 & 0.301                 & 0.543                 &                      & 1.247                   & 1.041                   & 0.635                     & -0.091                    & -0.497                    & -0.452                    & -0.442                    & -0.754                    & -0.687                           \\
0.2              & -0.671                & 0.671                 &                      & -0.097                 & 0.563                 & 0.420                  & -0.851                & -0.636                 & 0.122                 & 0.479                 &                      & 1.290                   & 0.997                   & 0.538                     & -0.213                    & -0.618                    & -0.575                    & -0.497                    & -0.590                    & -0.331                           \\
0.4              & -0.192                & 0.192                 &                      & -0.272                 & 0.332                 & 0.196                  & -0.568                & -0.374                 & 0.111                 & 0.575                 &                      & 0.507                   & 0.581                   & 0.472                     & -0.170                    & -0.336                    & -0.216                    & -0.138                    & -0.309                    & -0.392                           \\
0.6              & -0.176                & 0.176                 &                      & -0.419                 & 0.523                 & 0.212                  & -0.781                & -0.454                 & 0.013                 & 0.906                 &                      & 1.045                   & 0.491                   & 0.440                     & -0.371                    & -0.567                    & -0.341                    & -0.233                    & -0.318                    & -0.146                           \\
0.8              & -0.128                & 0.128                 &                      & -0.503                 & 0.347                 & 0.138                  & -0.557                & -0.274                 & -0.017                & 0.865                 &                      & 0.244                   & 0.253                   & 0.420                     & -0.274                    & -0.234                    & -0.097                    & 0.051                     & -0.104                    & -0.258                           \\
1.0             & -0.099                & 0.099                 &                      & -0.322                 & 0.245                 & 0.192                  & -0.382                & -0.130                 & -0.040                & 0.437                 &                      & 0.235                   & 0.171                   & 0.399                     & -0.220                    & -0.232                    & -0.066                    & -0.036                    & -0.047                    & -0.204                           \\ \hline
\end{tabular}
}
\label{tab:self_sim}
\end{table*}
\begin{table*}[ht!]
\centering
\caption{Details of the person preference score for OpenCLIP. Characters in each table are: 1) In gender: \textbf{M}an and \textbf{W}oman; 2) In ethnicity: \textbf{B}lack, \textbf{E}ast \textbf{A}sia, \textbf{I}ndian, \textbf{L}atino, \textbf{M}iddle \textbf{E}ast, \textbf{S}outheast \textbf{A}sia, and \textbf{Wh}ite.}
\renewcommand{\arraystretch}{1.6}
\resizebox{!}{52pt}{
\begin{tabular}{crrrrrrrrrrrrrrrrrrrr}
\hline
\multicolumn{1}{l}{} & \multicolumn{2}{c}{gender}                    & \multicolumn{1}{l}{} & \multicolumn{7}{c}{ethnicity}                                                                                                                                            & \multicolumn{1}{l}{} & \multicolumn{9}{c}{age}                                                                                                                                                                                                                                      \\ \cline{2-3} \cline{5-11} \cline{13-21} 
\multicolumn{1}{c}{\textbf{$\alpha$}} & \multicolumn{1}{c}{M} & \multicolumn{1}{c}{W} & \multicolumn{1}{l}{} & \multicolumn{1}{c}{EA} & \multicolumn{1}{c}{I} & \multicolumn{1}{c}{SA} & \multicolumn{1}{c}{Wh} & \multicolumn{1}{c}{ME} & \multicolumn{1}{c}{L} & \multicolumn{1}{c}{B} & \multicolumn{1}{l}{} & \multicolumn{1}{c}{0-2} & \multicolumn{1}{c}{3-9} & \multicolumn{1}{c}{10-19} & \multicolumn{1}{c}{20-29} & \multicolumn{1}{c}{30-39} & \multicolumn{1}{c}{40-49} & \multicolumn{1}{c}{50-59} & \multicolumn{1}{c}{60-69} & \multicolumn{1}{c}{more than 70} \\ \hline
0.0             & 0.646                 & 0.660                 &                      & 0.518                  & 0.486                 & 0.513                  & 0.507                 & 0.491                  & 0.505                 & 0.484                 &                      & 0.447                   & 0.430                   & 0.394                     & 0.380                     & 0.375                     & 0.365                     & 0.361                     & 0.357                     & 0.356                            \\
0.2              & 0.823                 & 0.895                 &                      & 0.709                  & 0.653                 & 0.668                  & 0.703                 & 0.674                  & 0.669                 & 0.621                 &                      & 0.851                   & 0.865                   & 0.896                     & 0.900                     & 0.895                     & 0.886                     & 0.883                     & 0.873                     & 0.859                            \\
0.4              & 0.904                 & 0.879                 &                      & 0.637                  & 0.581                 & 0.608                  & 0.641                 & 0.611                  & 0.602                 & 0.551                 &                      & 0.730                   & 0.758                   & 0.772                     & 0.791                     & 0.785                     & 0.770                     & 0.752                     & 0.726                     & 0.676                            \\
0.6              & 0.976                 & 0.994                 &                      & 0.353                  & 0.330                 & 0.329                  & 0.331                 & 0.304                  & 0.311                 & 0.307                 &                      & 0.825                   & 0.840                   & 0.838                     & 0.832                     & 0.808                     & 0.778                     & 0.741                     & 0.712                     & 0.702                            \\
0.8              & 0.974                 & 0.955                 &                      & 0.079                  & 0.055                 & 0.069                  & 0.089                 & 0.067                  & 0.068                 & 0.064                 &                      & 0.883                   & 0.951                   & 0.979                     & 0.984                     & 0.981                     & 0.978                     & 0.969                     & 0.952                     & 0.939                            \\
1.0             & 1.000                 & 1.000                 &                      & 0.149                  & 0.130                 & 0.134                  & 0.183                 & 0.164                  & 0.146                 & 0.113                 &                      & 0.943                   & 0.978                   & 0.988                     & 0.989                     & 0.985                     & 0.978                     & 0.962                     & 0.938                     & 0.905                            \\ \hline
\end{tabular}
}
\label{tab:person_preference}
\end{table*}

\begin{figure*}
  \centering
  \includegraphics[width = 0.98\textwidth]{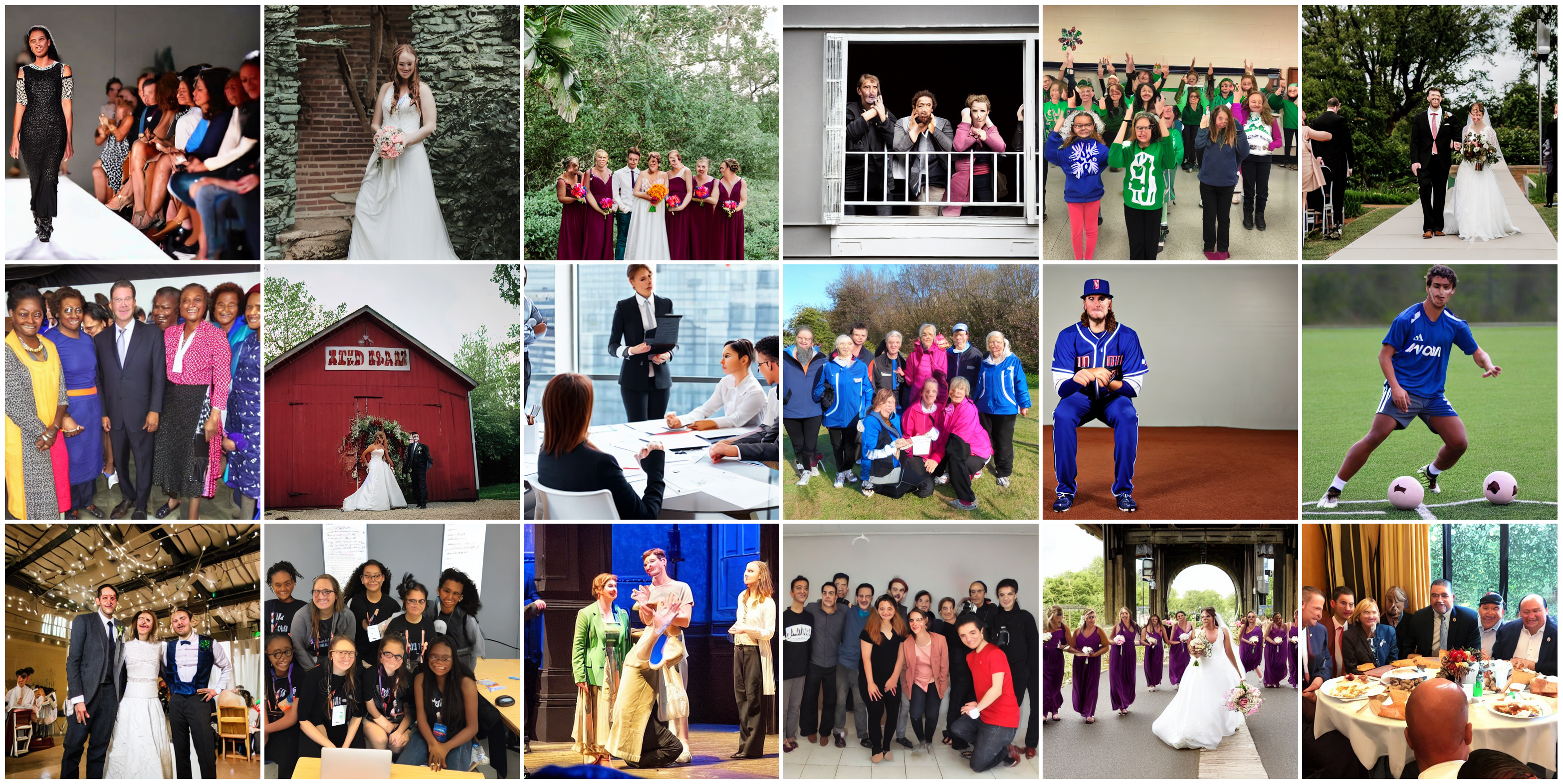}
  \caption{More examples of blurry faces in the generated images.}
  \label{fig:examples_blurry_face_more}
\end{figure*}

\begin{figure*}
  \centering
  \includegraphics[width = \textwidth]{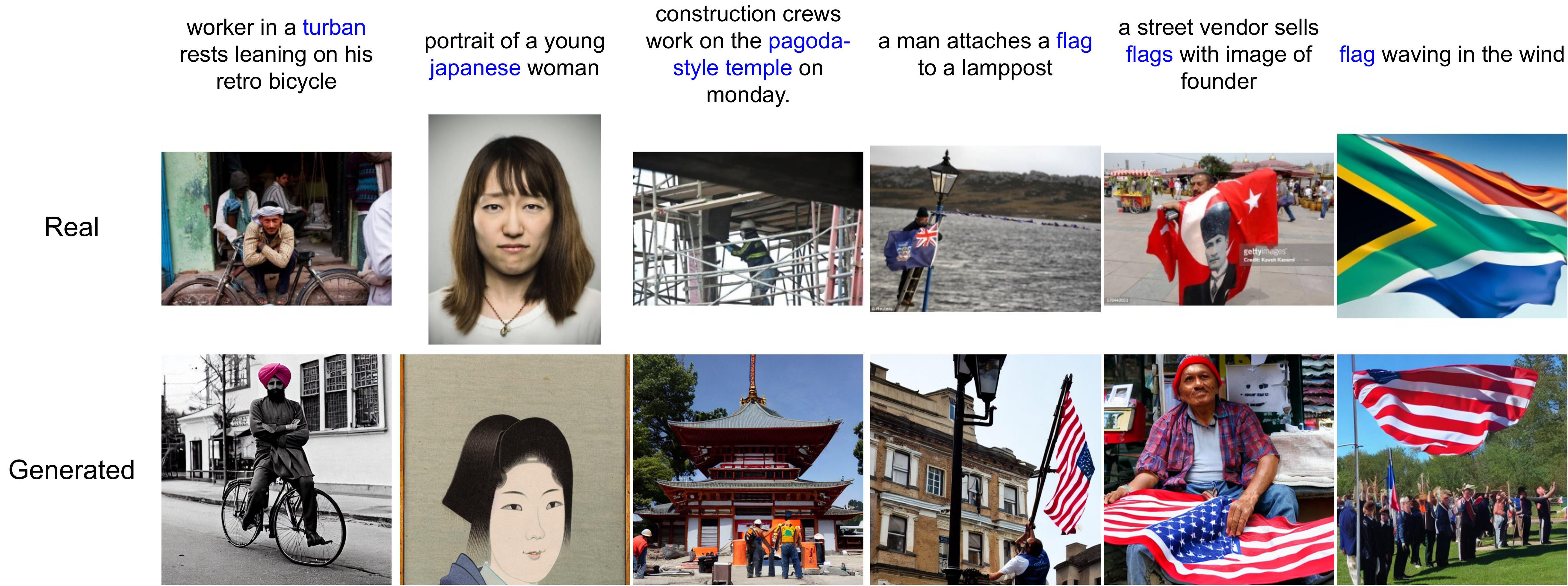}
  \caption{More examples of stereotyping in the generated images.}
  \label{fig:examples_overreact_more}
\end{figure*}

\end{document}